\documentclass[journal]{IEEEtran}
\usepackage{amsmath,amsfonts}
\usepackage{algorithmic}
\usepackage{algorithm}
\usepackage{array}
\usepackage{textcomp}
\usepackage{stfloats}
\usepackage{url}
\usepackage{cite}
\usepackage{verbatim}
\usepackage{graphicx}
\usepackage{balance}
\usepackage[hidelinks]{hyperref}
\usepackage{cleveref}
\usepackage{dsfont}
\usepackage{color}
\usepackage[caption=false,font=footnotesize]{subfig}
\usepackage{mathtools, cuted}

\usepackage{microtype}
\usepackage{multicol}
\usepackage{booktabs}

\usepackage{amsmath}
\usepackage{amssymb}
\usepackage{mathtools}
\usepackage{amsthm}

\theoremstyle{plain}
\newtheorem{theorem}{Theorem}
\newtheorem{proposition}[theorem]{Proposition}
\newtheorem{corollary}[theorem]{Corollary}
\theoremstyle{remark}
\newtheorem{property}[theorem]{Property}

\DeclareMathOperator{\E}{\mathbb{E}}

\newcommand{\train}{{\textsf{train}}}

\newcommand{\prior}{{\textsf{prior}}}
\newcommand{\model}{{\textsf{model}}}

\newcommand{\major}{{\textsf{major}}}
\newcommand{\minor}{{\textsf{minor}}}
\newcommand{\TLA}{{\textsf{TLA}}}
\newcommand{\GML}{{\textsf{GML}}}
\newcommand{\TW}{{\textsf{TWCE}}}
\newcommand{\softmax}{{\textsf{softmax}}}
\newcommand{\Bern}{{\textsf{Bern}}}
\newcommand{\Bin}{{\textsf{Bin}}}
\newcommand{\CE}{{\textsf{CE}}}
\newcommand{\maxx}{{\textsf{max}}}
\newcommand{\minn}{{\textsf{min}}}
\newcommand{\inter}{{\textsf{inter}}}
\newcommand{\intra}{{\textsf{intra}}}

\begin{document}
\title{Deep Minimax Classifiers for Imbalanced Datasets with a Small Number of Minority Samples}
\author{
  Hansung Choi and Daewon Seo
    \thanks{The authors are with the Electrical Engineering and Computer Science, DGIST, Daegu 42988, South Korea (e-mail: \{hansungchoi, dwseo\}@dgist.ac.kr).}
}

\maketitle

\begin{abstract}
The concept of a minimax classifier is well-established in statistical decision theory, but its implementation via neural networks remains challenging, particularly in scenarios with imbalanced training data having a limited number of samples for minority classes. To address this issue, we propose a novel minimax learning algorithm designed to minimize the risk of worst-performing classes. Our algorithm iterates through two steps: a minimization step that trains the model based on a selected target prior, and a maximization step that updates the target prior towards the adversarial prior for the trained model. In the minimization, we introduce a targeted logit-adjustment loss function that efficiently identifies optimal decision boundaries under the target prior. Moreover, based on a new prior-dependent generalization bound that we obtained, we theoretically prove that our loss function has a better generalization capability than existing loss functions. During the maximization, we refine the target prior by shifting it towards the adversarial prior, depending on the worst-performing classes rather than on per-class risk estimates. Our maximization method is particularly robust in the regime of a small number of samples. Additionally, to adapt to overparameterized neural networks, we partition the entire training dataset into two subsets: one for model training during the minimization step and the other for updating the target prior during the maximization step. Our proposed algorithm has a provable convergence property, and empirical results indicate that our algorithm performs better than or is comparable to existing methods. All codes are publicly available at https://github.com/hansung-choi/TLA-linear-ascent.
\end{abstract}

\begin{IEEEkeywords}
Minimax training, imbalanced data, adversarial prior
\end{IEEEkeywords}

\section{Introduction}

Classification using deep learning has become prevalent due to its ability to learn effective classification boundaries from training data. However, a common issue in many real-world applications is that a trained classifier results in risks that vary significantly across classes. Specifically, the classifier may perform poorly on test data if the test data's prior is significantly different from that of the training data. This issue is particularly prominent with class-imbalanced training data, where na\"{i}ve learning algorithms tend to be largely biased, overclassifying major classes and underclassifying minor ones~\cite{johnson2019survey, gilet2020discrete}. It means that, for instance of medical diagnosis, a vanilla-trained neural classifier is biased toward more frequent diseases rather than treating all diseases equally critical. 

When training with imbalanced data, a widely used goal is to improve class-balanced accuracy. To this end, it is common to modify loss functions to learn unbiased decision boundaries~\cite{du2023no,cui2019class, cao2019learning, menon2020long, ye2020identifying, kini2021label, fernando2021dynamically}. Early research introduced weight adjustment methods, where per-class weights are added to the cross-entropy (CE) loss function, assigning larger weights to minor classes and smaller weights to major classes~\cite{cui2019class,lin2017focal}. While these methods improve the balanced accuracy compared to vanilla CE loss, they suffer from damaged feature extraction~\cite{cao2019learning, ye2020identifying}, limited impact on moving decision boundaries~\cite{ byrd2019effect,menon2020long}, and unstability for highly imbalanced data~\cite{cui2019class}. To address these limitations, logit adjustment methods were developed~\cite{cao2019learning, menon2020long, ye2020identifying, kini2021label}. These approaches modify the CE loss by adding or multiplying per-class logits directly to the neural network outputs. They provide better control over moving decision boundaries~\cite{cao2019learning,menon2020long} and improved feature quality~\cite{cao2019learning, ye2020identifying}, leading to much higher balanced accuracy than the weight adjustment methods.

Another class of solutions is by augmenting the training data to effectively boost the representation of minor classes~\cite{ahn2023cuda, kandpal2023large, dablain2022deepsmote}. Improving the model's feature extraction through representation learning can also be an effective approach to the imbalanced data problem~\cite{yanneural,du2023global, li2023fcc, ma2022delving, peifeng2023feature,rangwani2024deit,lienhancing}. On the other side, semi-supervised approaches can generate sufficiently balanced data to solve imbalanced data problems~\cite{li2024twice, guo2022class}.

Since the loss modifications and other strategies mentioned above can be used together, recent studies have combined loss modification with their own proposed methods. For example, several works enhance feature extractors by incorporating previous loss modifications~\cite{rangwani2024deit, lienhancing, peifeng2023feature}. Other studies focus on boosting the overall effectiveness of previous loss functions using newly developed optimization techniques~\cite{rangwani2022escaping, zhou2023imbsam, luorevive, wang2023unified}, such as perturbing network parameters~\cite{rangwani2022escaping, zhou2023imbsam}. On the other hand, models can be enhanced by considering class-conditional likelihood~\cite{luorevive} or generalization bounds~\cite{wang2023unified}. In this way, previous research on imbalanced data has proposed effective loss functions and extra methods by considering other aspects of deep learning~\cite{rangwani2024deit, lienhancing, peifeng2023feature, rangwani2022escaping, zhou2023imbsam, luorevive, wang2023unified, li2024twice}.

However, most research with imbalanced data fails to increase the worst-class accuracy effectively since proposed loss functions~\cite{cui2019class,cao2019learning, menon2020long, ye2020identifying, kini2021label} rely solely on the size of the given training data without considering the accuracy of each class. For instance, consider a scenario where some minor classes are highly overlapped with each other while other minor classes are relatively separable in feature space. Then, the accuracies of highly overlapped minor classes are much lower than other minor classes. In this case, the effect of the size-based minority boosting may have a marginal effect on the highly overlapped minor classes. Due to this limitation, research for maximizing the worst-class accuracy is needed when maximizing the worst-class accuracy is an important goal or the true prior distribution of the test dataset largely deviates from the balanced distribution.

Minimax classifiers in statistical decision theory address this issue by minimizing the maximum class-conditional risk~\cite{VanTrees1968, Poor1994, fauss2021minimax}. This approach essentially yields a Bayesian classifier that minimizes the weighted class-conditional risks, with weights chosen by an adversary maximizing the weight-averaged Bayesian risk. Such a minimax classifier guarantees a certain level of class-conditional risk for all classes, making the model robust against changes in prior distributions.

Although the minimax classifier is well understood in theory, implementing it using neural networks with highly imbalanced data still remains challenging. Only a handful of results exist for this problem~\cite{li2023wat,alaiz2005minimax, alaiz2007minimax, sagawa2019distributionally, zhang2020coping, wei2023distributionally}, and they have significant limitations. For instance, some recent works~\cite{li2023wat,pethick2023revisiting,wei2023distributionally} require large balanced data in some parts of training, which is generally inaccessible in real-world imbalanced data problems. In addition, a minimax training algorithm, by its nature, alternates between a minimization step that finds the optimal model parameters for a fixed target prior and a maximization step that updates the target prior to be close to the adversarial prior while keeping the model parameters fixed. However, both steps in the current minimax training approach face critical issues when applied to imbalanced datasets. First, the existing weight adjustment method for the minimization step damages feature quality and has limited impact on moving decision boundaries~\cite{li2023wat,pethick2023revisiting,zhang2020coping, sagawa2019distributionally, alaiz2005minimax, alaiz2007minimax}. Second, the exponential gradient ascent (EGA) method, commonly used for the maximization step, may fail to find a true adversarial prior, particularly with a limited number of samples~\cite{li2023wat, wei2023distributionally, zhang2020coping, sagawa2019distributionally, alaiz2005minimax, alaiz2007minimax}. As a result, existing minimax training solutions exhibit limited performance on imbalanced datasets with a small number of minority samples. Detailed theoretical and experimental justifications are provided in Sections~\ref{subsec:TLA method}, \ref{sec:linear ascent}, and \ref{sec:minmax_result}.

To solve the issues of the current minimax training, we propose a new targeted logit adjustment (TLA) loss function for the minimization step, and the linear ascent method for the maximization step. Our proposed method has provable convergence properties to the minimax solution and is robust with respect to a small number of minor class samples. Also, our method achieves the highest worst class accuracy compared to the previous minimax training and imbalanced data research. The following summarizes our main contributions.

\begin{itemize}
    \item Our TLA loss function for the minimization step is newly designed to effectively enhance the accuracy under a given target prior. Specifically, the logit terms of the CE loss are adjusted based on the target prior. To justify it, we analytically derive a new prior-dependent generalization bound, by which we show that the model trained with our TLA loss has a better generalization ability than the model trained with the previous weighting methods under the adversarial prior. In addition, we empirically demonstrate that the model trained with our TLA loss successfully moves decision boundaries to the desired locations and does not damage feature quality.

    \item To update the target prior close to the adversarial one, we propose a new linear ascent algorithm that shifts the target prior towards the direction of the worst-performing classes. Unlike existing algorithms that adjust a prior in a ``soft'' manner based on per-class risks, our ascent algorithm depends only on finding the correct worst class. As a result, our proposed algorithm is more robust especially when the number of minority samples is small, which is not only experimentally justified but also theoretically derived.
    
    \item To address the zero-error problem that arises in the training of overparameterized neural networks~\cite{sagawa2019distributionally}, we partition the training data into two sets: one for model training and the other for updating the target prior. Once the target prior is updated to match the adversarial prior, two subsets are merged again, on which the model is fine-tuned.
\end{itemize}

The remaining part of this paper is organized as follows. In Section~\ref{sec:related_work}, our problem is formally defined, and related works are reviewed. In Section~\ref{sec:proposed method}, our algorithm is discussed. In Section~\ref{sec:experiment}, experiment results on standard image datasets are presented. Finally, Section~\ref{sec:conclusion} concludes the paper.

\section{Problem Definition and Related Work} \label{sec:related_work}

We consider a supervised classification problem with given training data, where each training sample consists of an instance $x \in \mathcal{X} \subset \mathbb{R}^d$ and a class $y \in [K] := \{1, ..., K\}$. The training data follow the distribution $p^\train(x,y) = \pi^{\train}_{y} p(x|y)$ where $\pi^\train_y := \pi^\train(y)$ is the prior distribution of the training data (i.e., the distribution of classes), and $p(x|y)$ is a fixed class-conditional distribution that remains unchanged for the test data. Motivated by the distributional shift in test data's prior, commonly observed in imbalanced data, our goal is to minimize the risk of the worst-performing class in test data. Specifically, let $R(\pi, \theta)$ be the total $0$-$1$ loss (i.e., total error probability) measured by test data having prior $\pi$, not necessarily identical to $\pi^\train$, at model parameters $\theta$. In other words, $R(\pi, \theta) = \sum_{y=1}^{K} \pi_{y} P^{(e)}_{y,\theta}$, where $P^{(e)}_{y,\theta}$ is the probability of error evaluated at model parameters $\theta$ for class $y$. Then, our objective is to find the model parameters $\theta^*$ that achieves
\begin{align}
    R^* := \min_{\theta} \max_{\pi} R(\pi,\theta). \label{eq:minmax_formulation}
\end{align}

In classic decision theory, it is known that the Bayes-optimal decision rule for the least-favorable prior is the minimax decision rule~\cite{Poor1994, fauss2021minimax}. However, solving \eqref{eq:minmax_formulation} by an alternating optimization algorithm does not always give a convergent solution. To deal with this, a common relaxation is by the minimax inequality~\cite[Chapter~7]{deisenroth2020mathematics}: For any $R(\cdot, \cdot)$,
\begin{align}
    \min_{\theta} \max_{\pi} R(\pi,\theta) \ge \max_{\pi} \min_{\theta} R(\pi,\theta) =: \max_{\pi} R(\pi,\theta^*(\pi)),
\end{align}
where $\theta^*(\pi) := \arg\min_{\theta} R(\pi,\theta)$.\footnote{By von Neumann's minimax theorem~\cite[Chapter~6]{aubin2006applied}, the equality holds if $R(\pi, \theta)$ is concave in $\pi$ and convex in $\theta$, and the spaces for $\pi, \theta$ are compact convex sets.} Note that since $R(\pi, \theta^*(\pi))$ is concave in $\pi$~\cite{gilet2020discrete}, it can be readily maximized. Similarly, the inner minimization can be solved by neural networks. Therefore, instead of \eqref{eq:minmax_formulation}, our new objective is 
\begin{align}
    R^{\dagger} := \max_{\pi} \min_{\theta} R(\pi,\theta). \label{eq:maxmin_formulation}
\end{align}

The process solving \eqref{eq:maxmin_formulation} can be broken down into two steps: the inner minimization that finds the best classifier $\theta$ for a given target prior $\pi$, and the outer maximization that finds an adversarial prior for a trained classifier. In the following, we separately review how to solve the inner minimization and the outer maximization using neural networks and point out the drawbacks of existing methods.

\subsection{Related Work for Inner Minimization} \label{subsec:related_work_min}

To learn a model for $\pi \ne \pi^{\train}$ from given training data, importance weighting is widely used ~\cite{li2023wat, zhang2020coping, sagawa2019distributionally, alaiz2005minimax, alaiz2007minimax}. For a loss function $l(y,f(x))$ where $f(x)$ is a model output, a different per-class weight is assigned to each class to compensate for the disparity between $\pi$ and $\pi^{\train}$. Specifically, to minimize the loss at $\pi \ne \pi^{\train}$, one can assign weights $\pi_{y} / \pi^{\train}_{y}$ for each class and then train a network on the training data. Then,
\begin{align}
&\E_{(x,y)\sim \pi^{\train}_{y} p(x|y)} \left[ \frac{\pi_{y}}{\pi^{\train}_{y}} l(y,f(x))  \right] \\
&= \E_{(x,y) \sim \pi_{y} p(x|y)} \left[ l(y,f(x))  \right].
\end{align}
Thus, minimizing a new loss $\frac{\pi_{y}}{\pi^{\train}_{y}} l(y,f(x))$ on training data is the same as minimizing $l(y,f(x))$ on a new dataset with prior $\pi$.

However, this weight adjustment method turns out to damage the quality of extracted features and have a limited effect on moving decision boundaries for overparametrized networks~\cite{cao2019learning, menon2020long, ye2020identifying, kini2021label}. To address them, we propose a targeted logit adjustment (TLA) loss function. The model trained with our proposed loss successfully increases the accuracy of the worst classes by finding good decision boundaries under a given adversarial prior distribution without damaging feature quality. 
 Moreover, based on our newly derived prior-dependent generalization bound, we theoretically prove that the model trained with our TLA loss has a better generalization ability than the model with the previous weighting methods under the adversarial prior distribution. Detailed theoretical analysis and experimental validations are given in Sections~\ref{subsec:TLA method} and~\ref{sec:minmax_result}.

\subsection{Related Work for Outer Maximization} \label{subsec:related_work_max}

Most of the literature relies on exponential gradient ascent (EGA) to find an adversarial prior $\pi^* := \arg\max_{\pi} R(\pi,\theta^*(\pi))$~\cite{li2023wat, wei2023distributionally, zhang2020coping, sagawa2019distributionally, alaiz2005minimax, alaiz2007minimax}. The standard EGA updates $\pi$ so that it becomes close to $\pi^*$ based on class-conditional $0$-$1$ loss $P^{(e)}_{y,\theta}$ with step size $\alpha$. To be precise, letting $\pi^t$ be the prior found at the $t$-th iteration, the EGA computes
\begin{align}
\pi^{t+1}_{y} &\leftarrow \frac{\pi^t_{y} \exp (\alpha P^{(e)}_{y, \theta})}{\sum_{y'=1}^{K} \pi^{t}_{y'} \exp (\alpha P^{(e)}_{y',\theta})}.
\end{align}
If $P^{(e)}_{y, \theta}$ is computed accurately, then the updated prior converges to the adversarial prior $\pi^*$.

However, in practice, since the exact class-conditional risks are unavailable, the EGA is based on an estimate 
\begin{align}
    \widehat{P}^{(e)}_{y,\theta}=\frac{1}{N_y}\sum_{y_n=y}\mathds{1}\left[\widehat{y}_n(x_n;\theta) \ne y \right],
    \label{error estimate}
\end{align}
where $N_y$ is the number of training samples for class $y$, $\widehat{y}_n(x_n;\theta)$ is the predicted class for given instance $x_n$ with $\theta$, and $\mathds{1}[\cdot]$ is the indicator function. In other words, the prior is updated based on ``soft'' estimate of risks. Hence, when the estimate is more inaccurate, the updated prior converges to a more incorrect prior or oscillates. This issue arises in particular when the number of minority samples is small because
\begin{align}
    \E\left[\left(\widehat{P}^{(e)}_{y,\theta}-P^{(e)}_{y,\theta}\right)^2\right] \propto \frac{1}{N_y}
    \label{MSEerror}
\end{align}
if samples are independent and identically distributed. Hence, with small minority samples, the estimated risks are typically highly inaccurate, and thus, the updated prior may not converge to $\pi^*= \arg\max_{\pi} R(\pi,\theta^*(\pi))$. To overcome this limitation, we propose a linear ascent method. Our linear ascent finds $\pi^*= \arg\max_{\pi} R(\pi,\theta^*(\pi))$ more efficiently even with a smaller number of per-class samples.

\section{Our Proposed Method} \label{sec:proposed method}
Note that solving $R^\dagger = \max_{\pi} \min_{\theta} R(\pi,\theta)$ requires two steps: 1) finding the Bayes-optimal classifier for a given target prior, and 2) finding the adversarial prior for a given model. To address these two, we first introduce a new loss function, referred to as the targeted logit adjustment (TLA) loss, for the inner minimization. The TLA loss enhances the accuracy of a model under a given target prior. Secondly, we present a novel ascent algorithm for the outer maximization that efficiently updates the target prior close to an adversarial prior, even with a small number of samples. In addition, we implement Algorithm~\ref{alg:minmax}, which properly integrates the preceding two steps by partitioning the training samples into two sets.

\subsection{Targeted Logit Adjustment Loss for Model Training} \label{subsec:TLA method}

As reviewed in Section~\ref{subsec:related_work_min}, adjusting weights based on the adversarial prior suffers from poor feature extraction and vanishing effect for overparameterized networks. Our targeted logit adjustment (TLA) loss that mitigates the aforementioned problems is defined as
\begin{align}
l_{\TLA}(y,f(x) ; \pi^t) = -\log\left( \frac{\exp( f_{y}(x)+ \ell_y(\pi^t) )}{\sum_{y'} \exp( f_{y'}(x) +  \ell_{y'}(\pi^t) )} \right),
\end{align}
where $\ell_y (\pi^t) = \tau (\log \pi^{\train}_{y} - \log \pi^t_{y})$ with constant $\tau>0$ and $f_{y}(x)$ is the $y$-th output of the neural network. A neural network trained with the TLA loss minimizes $R(\pi^t,\theta)$ due to $\ell_y (\pi^t)$. To have the TLA loss small, $\exp( f_{y}(x)+ \ell_y(\pi^t) )$ should be large. For large $\pi^t_{y}$, $\ell_y (\pi^t)$ is small. Thus, for large $\pi^t_{y}$, $f_{y}(x)$ should be large to have large $\exp( f_{y}(x)+ \ell_y(\pi^t) )$, leading to small TLA loss. Note that large $f_{y}(x)$ for large $\pi^t_{y}$ leads high accuracy under a target prior $\pi^t$. Thus, training with the TLA loss enhances the accuracy of the model under a target prior $\pi^t$. The detailed effect of $\ell_y (\pi^t)$ in TLA loss and its analysis are described as follows.

\medskip
\noindent \textbf{Comparison with Weighting Method:} To illustrate the advantage of the TLA loss over weight-based methods, consider the following weight-adjusted CE loss~\cite{zhang2020coping, sagawa2019distributionally, alaiz2005minimax, alaiz2007minimax}:
\begin{align}
l_{\TW}(y,f(x) ; \pi^t) &= -\frac{\pi^t_{y}}{\pi_{y}^{\train}}  \log\left(\frac{\exp( f_{y}(x) )}{\sum_{y'}\exp( f_{y'}(x) )}\right).
\end{align}
We denote this loss as a targeted-weight CE (TWCE) loss to distinguish it from the general weighted CE loss. The TWCE loss is designed to increase the weighted accuracy according to the target prior. However, it is known that well-trained overparameterized networks perfectly fit the training data~\cite{zhang2017understanding,byrd2019effect}, which implies that the logarithmic term converges to near $0$~\cite{papyan2020prevalence}, leading to the entire TWCE loss close to zero no matter what $\pi^t_{y} / \pi_{y}^{\train}$ is. It in turn implies that the model is not effectively optimized.

Unlike weight terms of $l_{\TW}(y,f(x) ; \pi^t)$, logit terms of $l_{\TLA}(y,f(x) ; \pi^t)$ directly adjust neural network output by $\tau (\log \pi^{\train}_{y} - \log \pi^t_{y})$, which directly affects per-class decision boundaries and fixes the vanishing effect of the TWCE loss. Suppose that class $y$ is the worst class and the adversarial prior correctly over-weights it, i.e., $\pi^{\train}_{y} < \pi^t_{y}$. Then, a model will increase $f_y(x)$ more than $|\tau (\log \pi^{\train}_{y} - \log \pi^t_{y})|$ to minimize the TLA loss. As a result, the model will be more likely to predict $y$, which implies that the model effectively moves its decision boundary further away from class $y$, i.e., the accuracy for class $y$ increases.

\medskip
\noindent \textbf{Analysis:} To show the effect of $l_{\TLA}(y,f(x) ; \pi^t)$ for inner minimization step, note that for each instance $x$, the Bayes-optimal classifier predicts $y$ by declaring the posterior distribution $p(y|x)$ of $p(x,y) = \pi_y p(x|y)$. As the network's output is converted to $p(y|x)$ by softmax normalization, the network's decision is Bayes optimal at $\pi$ if $\exp\left(f_{y}(x)\right) \propto p(y|x)$. The goal of the inner minimization is to minimize $R(\pi, \theta)$ for a given $\pi$ that is possibly different from the training data's prior $\pi^\train$. The following proposition shows that this can be done by properly manipulating the output of $\pi^\train$-trained neural networks.

\begin{proposition} \label{pro:logit}
Consider the Bayes-optimal model for training data such that $\exp\left(f_{y}^{\train}(x)\right) \propto p^{\train}(y|x)$ and another Bayes-optimal model for data with another prior $\pi$ such that $\exp\left(f_{y}(x)\right) \propto p(y|x)$. Then, for some $c(x)$,
\begin{align}
    f_{y}^{\train}(x) = f_{y}(x) + \left( \log \pi^{\train}_{y} - \log \pi_{y}  + \log c(x) \right).
\end{align}
\end{proposition}
\begin{IEEEproof}
As $p(y|x) = \frac{p(x|y) \pi_{y}}{p(x)} = \frac{p(x|y) \pi_{y}}{\sum_{y'} p(x|y') \pi_{y'}}$,
\begin{align}
p^{\train}(y|x) &= p(x|y)\frac{\pi^{\train}_{y}}{p^{\train}(x)} = p(y|x) \frac{p(x)}{\pi_{y}} \frac{\pi^{\train}_{y}}{p^{\train}(x)}. \label{eq:pdf_conversion}
\end{align}
Also, the assumption $\exp(f_y(x)) \propto p(y|x)$ implies that $\frac{\exp\left(f_{y}(x)\right)}{\sum_{y'} \exp\left(f_{y'}(x)\right)}=p(y|x)$. Hence, \eqref{eq:pdf_conversion} can be written using the network's output as follows.
\begin{align}
&\frac{\exp\left(f_{y}^{\train}(x)\right)}{\sum_{y'} \exp\left(f_{y'}^{\train}(x)\right)} =\frac{\exp\left(f_{y}(x)\right)}{\sum_{y'} \exp\left(f_{y'}(x)\right)} \frac{p(x) \pi^{\train}_{y}}{\pi_y p^{\train}(x)}.
\end{align}
Thus, it holds that $\exp\left(f_{y}^{\train}(x)\right) = c(x) \exp\left(f_{y}(x)\right) \frac{\pi^{\train}_{y}}{\pi_{y}}$ with proper $c(x)$, or equivalently, 
\begin{align}
f_{y}^{\train}(x) = f_{y}(x) + \left( \log \pi^{\train}_{y} - \log \pi_{y}  + \log c(x) \right).
\end{align}
\end{IEEEproof}

In Proposition~\ref{pro:logit}, we have seen that $\min_{\theta} R(\pi, \theta)$ at an arbitrary $\pi$ can be achieved by adjusting the output of a well-trained model on the training data. The following theorem shows that training with our loss function, the TLA loss with $\tau=1$, finds such an adjusted output of the neural network that is optimal for any given $\pi$.

\begin{theorem} \label{thm:robust logit loss}
Suppose that the neural network trained on the training data with the vanilla CE loss outputs $f_y(x) = f_y^\train(x)$ such that $\exp\left(f_{y}^{\train}(x)\right) \propto p^{\train}(y|x)$. Then, for a target prior $\pi^t$, the neural network trained on the training data with the TLA loss with $\tau=1$ outputs $f_y(x) = f_y^t(x)$ such that 
\begin{align}
    \exp\left(f_{y}^{t}(x)\right) \propto p^t(y|x) = \frac{ \pi^t_{y} p(x|y) }{ \sum_{y'} \pi^t_{y'} p(x|y') }.
\end{align}
\end{theorem}

\begin{IEEEproof}
Recall that the vanilla CE loss function is
\begin{align}
    l_{\CE}(y,f(x)) = -\log\left(\frac{\exp( f_{y}(x) )}{\sum_{y'}\exp( f_{y'}(x) )}\right), \label{eq:vanilla_CE}
\end{align}
by which the trained network on data with prior $\pi^\train$ finds $f_y(x) = f_y^\train(x)$ such that $\exp\left(f_{y}^{\train}(x)\right) \propto p^{\train}(y|x)$. Hence, replacing $f_y(x)$ term in \eqref{eq:vanilla_CE} with
\begin{align}
    f_{y}(x) + \left( \log \pi^{\train}_{y} - \log \pi^{t}_{y} + \log c(x) \right)
\end{align}
modifies the output $f_y(x)$ so that
\begin{align}
    f_{y}(x) + \left( \log \pi^{\train}_{y} - \log \pi^{t}_{y} + \log c(x) \right) = f_{y}^{\train}(x).
\end{align}
Since it satisfies the relation in Proposition~\ref{pro:logit}, the trained model with the TLA loss is optimal at prior $\pi^t$. Finally, note that omitting $\log c(x)$ term does not change the training process as $e^{\log c(x)} = c(x)$ terms in the numerator and the denominator cancel each other. This results in the bias in the TLA loss $\ell_y(\pi^t) = \log \pi^\train_{y} - \log \pi^{t}_{y}$.
\end{IEEEproof}

The assumption of Theorem~\ref{thm:robust logit loss} is widely used in deep learning literature~\cite{menon2020long,zhang2020bayes}. In the community, it is now standard to assume that neural networks with cross-entropy loss find Bayes-optimal decision boundaries under ideal conditions, such as being well-trained and having a large number of samples~\cite{pires2016multiclass}. There is extensive work supporting this phenomenon~\cite{zhang2020bayes,awasthi2022h,mao2023structured}, as well as methods for creating Bayes-optimal models~\cite{radhakrishnan2023wide}. Based on these concrete backgrounds, a model trained with our TLA loss (under the ideal condition) is expected to make output $\exp\left(f_{y}^{t}(x)\right) \propto p^t(y|x)$.

When the number of training data is sufficient, setting $\tau=1$ is enough to make the model trained with our TLA loss converge to the Bayes-optimal model for target prior $\pi^t$. However, when the number is not sufficient, it is often not sufficient to move decision boundaries as per bias term $\log \pi^{\train}_{y} - \log \pi^{t}_{y}$. To this end, we introduce an extra factor $\tau$ that controls the bias more effectively, leading to moving decision boundaries more effectively. As demonstrated in Section~\ref{sec:experiment}, unlike the TWCE loss function, the model trained with our TLA loss effectively adjusts decision boundaries without compromising feature quality, leading to the best worst-class accuracy among all methods.

This effectiveness of TLA loss over other losses can also be supported by our new prior-dependent generalization bound, which is an adapted version of the recent generalization bound~\cite{wang2023unified}. The bound indicates that training with the TLA loss results in a tighter test loss bound under any prior than the TWCE loss, i.e., as the target prior is adversarial, it achieves a smaller error for the worst-performing class.

To discuss the bound, we introduce canonical definitions for Rademacher-based generalization bounds. Let $\hat{L}^{l}(f), L^{l}_{\pi}(f)$ respectively be empirical (under the training data) and expected (under an arbitrary prior $\pi$) losses for individual loss $l(\cdot, \cdot)$:
\begin{align}
\hat{L}^{l}(f) &:= \frac{1}{N} \sum_{(x_n,y_n) \in D_{\train}} l(y_n,f(x_n)), \\
L^{l}_{\pi}(f) &:= \E_{(x,y)\sim\pi_y p(x|y)}\left[ l(y,f(x)) \right].
\end{align}
Note that $\hat{L}^{l}(f) \approx 0$ in the overparametrized regime, but $L^{l}_{\pi}(f) > 0$ in general. Also, letting $\mathcal{F}$ be the collection of neural network functions, define $\mathcal{G}^l := \{ l \circ f: f \in \mathcal{F} \}$ be the set of composite loss functions $l(y,f(x))$. Then, the empirical Rademacher complexity $\hat{\mathfrak{R}}_N(\mathcal{G}^l)$ is defined as follows.
\begin{align}
    \hat{\mathfrak{R}}_N(\mathcal{G}^l) := \E_{\sigma}\left[ \sup_{g \in \mathcal{G}^l} \frac{1}{N} \sum_{n=1}^N \sigma_n g(z_n) \right],
\end{align}
where $\sigma_n$ is i.i.d.~Rademacher random variable and $z_n := (x_n,y_n)$.

With these definitions on one hand, we assume that a loss function $l(y, f(x))$ can be written in general form
\begin{align}
l(y,f(x)) &= -w_{y} \log \left( \frac{ \exp (\Delta_{y} f_{y}(x) + \ell_{y})}{\sum_{y' \in [K]} \exp (\Delta_{y'} f_{y'}(x) + \ell_{y'})} \right)
\label{VS_loss}
\end{align}
where per-class weight $w_{y}$, per-class multiplicative logit $\Delta_y$, and per-class additive logit $\ell_y$ are tunable hyperparameters. Then, a prior-dependent generalization bound is characterized as follows.\footnote{ For analytical tractability, it is standard to assume that the loss is bounded and Lipschitz continuous, e.g.~\cite{wang2023unified}. This holds true when the data are standardized and neural networks are regularized properly. } The proof is provided in Appendix~\ref{app:gen2}.
\begin{theorem} \label{thm:gen2}
Suppose a loss function $l(y, f(x))$ of interest is bounded by $B > 0$, i.e., $0 \le l(y,f(x)) \le B$, Lipschitz continuous, and the above assumption. Also, let $\pi_{\maxx}, \pi^\train_{\minn}$ be the priors of the largest class in the target prior and the smallest classes in training data, respectively. Then, for any given $0 < \delta < 1$ and training data $D_\train$, with probability at least $1-\delta$, the following bound holds
\begin{align}
     L^{l}_{\pi}(f) \le \Phi + \frac{\pi_{\maxx}}{\pi^{\train}_{\minn}} \cdot \hat{\mathfrak{R}}_N(\mathcal{F}) \cdot \sum_{y=1}^K w_y \overline{\Delta_y} \sqrt{\pi^{\train}_y} \Psi_y,
\end{align}
where
\begin{align}
    \Phi &= \Phi(l,\delta,\pi) := \frac{\pi_{\maxx}}{\pi^{\train}_{\minn}} \left[ \hat{L}^{l}(f) + 3B \sqrt{\frac{\log{2/\delta}}{2N}} \right], \\
    \overline{\Delta_y} &:= \sqrt{\Delta^2_y + \bigg( \sum_{y' \ne y} \Delta_{y'} \bigg)^2}, \\
    \Psi_y &= \Psi_y(l,f) := 1-\softmax \left(\Delta_y \min_{x \sim p(x|y)} f_y(x) + \ell_y \right).
\end{align}
\end{theorem}
First note that $\hat{L}^{l}(f)$ in $\Phi$ vanishes in well-trained overparametrized neural networks, and the remaining term in $\Phi$ is a constant. Then, evaluating the bound at our TLA and the previous TWCE loss and letting $S_y(f):=\min_{x\sim p(x|y)} f_y(x)$ for brevity give the following generalization bounds at an adversarial prior distribution $\pi^*$:
\begin{align}
    &L^{l_{\TLA}}_{\pi^*}(f) \le \text{constant} + \\
    &~ \frac{\pi^*_{\maxx}}{\pi^{\train}_{\minn}}\hat{\mathfrak{R}}_N(\mathcal{F}) \sum_{y=1}^K \sqrt{K^2-2K+2} \sqrt{\pi^{\train}_y} \Psi_y(l_{\TLA},f), \\
    &L^{l_{\TW}}_{\pi^*}(f) \le \text{constant} + \\
    &~ \frac{\pi^*_{\maxx}}{\pi^{\train}_{\minn}}\hat{\mathfrak{R}}_N(\mathcal{F}) \sum_{y=1}^K \sqrt{K^2-2K+2} \frac{\pi^*_y}{\sqrt{\pi^{\train}_y}}\Psi_y(l_{\TW},f),
\end{align}
where
\begin{align}
     \Psi_y(l_{\TLA},f) &=  1-\softmax(S_y(f) + \tau (\log \pi^{\train}_{y} - \log \pi^*_{y})), \\
     \Psi_y(l_{\TW},f) &= 1-\softmax(S_y(f)).
\end{align}

As one can see, the only difference is by terms $\sqrt{\pi^{\train}_y} \Psi_y(l_{\TLA},f)$ for the TLA loss and $\frac{\pi^*_y}{\sqrt{\pi^{\train}_y}}\Psi_y(l_{\TW},f)$ for the TWCE loss. For these terms, one important property can be observed: For the worst-performing class $y$ with a limited number of samples, $\frac{\pi^*_y}{\sqrt{\pi^{\train}_y}}$ is typically much larger than $1$ due to the ``adversarial'' nature of $\pi^*$,\footnote{ To maximize the $0$-$1$ total risk $R(\pi,\theta)$, adversarial prior $\pi^*$ should maximize the prior value of the worst class as much as possible, where the worst class is one of the minor classes due to its small sample size. } while $\sqrt{\pi^{\train}_y}$, $\Psi_y(l_{\TLA},f)$, and $\Psi_y(l_{\TW},f)$ are all bounded by $1$. This property concludes that our TLA loss function shows much smaller generalization bound than that of the TWCE loss. Since training error is nearly zero in overparametrized regime and $\pi^*$ is adversarial, it implies that the TLA loss results in a smaller test error for the worst-performing class.

\subsection{Linear Ascent for Finding Adversarial Prior} \label{sec:linear ascent}

The widely used exponential gradient ascent (EGA)~\cite{li2023wat, wei2023distributionally, zhang2020coping, sagawa2019distributionally, alaiz2005minimax, alaiz2007minimax} that finds an adversarial prior often does not converge to the correct $\pi^*= \arg\max_{\pi} R(\pi,\theta^*(\pi))$, especially when the number of per-class samples is not sufficiently large. This is because the ``soft'' estimates of class-conditional risks that the EGA relies on are inaccurate when there are a small number of samples. To alleviate this issue, we propose a new ascent update method, referred to as linear ascent, that determines the direction of the ascent relying only on whether or not each class belongs to the $M$ worst classes, i.e., ``hard'' estimate. Let $\theta^t$ be the optimal model parameters for a fixed $\pi^t$, i.e., $\theta^t = \arg\min_{\theta}R(\pi^t,\theta)$. Also, let $\pi^{t,\maxx,M}$ be the normalized binary vector that indicates the $M$ worst classes, i.e.,
\begin{align}
\label{eq:linear_direction}
    \pi_y^{t,\maxx,M} = \begin{cases}
        \frac{1}{M} & \text{if $y$ belongs to the $M$ worst classes at $\theta^t$,} \\
        0 & \text{otherwise.}
    \end{cases}
\end{align}
Then, our linear ascent algorithm updates the prior as
\begin{align}
\pi^{t+1} &\leftarrow \pi^t + \alpha (\pi^{t,\maxx,M} - \pi^t),
\end{align}
where $\alpha$ is a step size such that $0<\alpha<1$. As can be seen, this ascent method shifts prior to the linear direction of the $M$ worst-performing classes; hence, it is referred to as linear ascent. It increases the prior values of the $M$ worst classes and decreases those of the others. We set $M$ such that each of $M$ worst classes has a similar error probability to that of the worst class. If $M$ is bigger, the worst class is more likely included in the $M$ worst classes, while the value of $\pi^t$ moves by only a small distance since $0 \le \pi_y^{t,\maxx,M} \le \frac{1}{M}$. Thus, $M$ should be set as small as possible under the constraint of finding the worst class correctly. 

\begin{table}[t]
\centering
\caption{Accuracy results for CIFAR10 step-imbalanced data with imbalance ratio $\rho=0.01$. The accuracy of the worst class is significantly lower than that of the other classes.}
\label{tab:CIFAR10_0.01_step}
\begin{tabular}{|l|l|l|l|}
\hline
Loss & Worst class & Minority & Balanced \\ \hline
CE                      & $3.16$  & $17.94$   & $56.83$   \\ \hline
LA~\cite{menon2020long} & $41.72$ & $60.90$   & $72.22$   \\ \hline
VS~\cite{kini2021label} & $33.62$ & $62.01$   & $70.63$   \\ \hline
\end{tabular}
\vspace{-0.1in}
\end{table}

\begin{figure*}[hb]
\hrule
\begin{align}
\label{eq:estimated as m-th worst}
&\sum_{y_1 \ne 1}\sum_{y_2 \ne 1,y_1}...\sum_{y_{m-1} \ne 1,y_1,...,y_{m-2}} \left[ \prod_{y' \ne 1,y_1,...,y_{m-1}} \Pr\left[\hat{P}_{1, \theta}^{(e)} > \hat{P}_{y', \theta}^{(e)}\right]  \right ] \left[ \prod_{y' \in \{y_1,...,y_{m-1}\}} \Pr\left[\hat{P}_{1, \theta}^{(e)} \le \hat{P}_{y', \theta}^{(e)}\right]  \right ],
\end{align}
where
\begin{align}
&\Pr\left[\hat{P}_{y, \theta}^{(e)} > \hat{P}_{y', \theta}^{(e)}\right]=\sum_{i=0}^{N-1} \sum_{n=1}^{N-i} \Bin\left( i+n,N,P_{y, \theta}^{(e)} \right) \Bin\left(i,N,P_{y', \theta}^{(e)} \right),\\
&\Pr\left[\hat{P}_{y, \theta}^{(e)} \le \hat{P}_{y', \theta}^{(e)}\right]=\sum_{n=0}^{N} \sum_{i=0}^{N-n} \Bin\left(n,N,P_{y, \theta}^{(e)} \right) \Bin\left(n+i,N,P_{y', \theta}^{(e)} \right)
\end{align}
with $\Bin\left( n, N, P \right)$ being the binomial distribution for the number of success $n$, the number of trials $N$, and the success probability of each trial $P$.
\end{figure*}

\noindent \textbf{Comparison with EGA:} Before discussing the EGA, note that when a dataset is highly imbalanced, and some classes have very fewer samples than others, a trained model frequently shows the performance such that several worst classes have much lower accuracies than the others. For instance, Table~\ref{tab:CIFAR10_0.01_step} shows the performance of several existing training methods for CIFAR10 dataset, the setting for which is detailed in Section~\ref{sec:experiment}. One can see that for all methods, the worst-class accuracy is even much lower than that of the other minor classes. 

In such a scenario, finding which class is the worst is highly accurate, while an empirical estimate $\hat{P}_{y, \theta}^{(e)}$ for minor class is highly inaccurate due to \eqref{MSEerror}. To update $\pi^t$ with the desired true ascent direction, our linear ascent method needs to find the true worst class, while EGA needs to estimate $\exp{( \hat{P}_{y, \theta}^{(e)} )}$ accurately as possible. In this part, we show that the error probability of finding the worst class is more robust than the MSE of $\exp{( \hat{P}_{y, \theta}^{(e)} )}$ when the number of samples is small, which means that our linear ascent method is more robust than the EGA method.

To show the robustness of our method, we derive and compare the error probability of finding the worst class and the MSE of $\exp{( \hat{P}_{y, \theta}^{(e)} )}$ with respect to the number of samples $N$. First, the accuracy of estimating the worst class is stated in the next theorem, the proof of which is in Appendix~\ref{app:find the correct worst class}.
\begin{theorem}
\label{thm:find the correct worst class}
Suppose that the true class-wise error probabilities are ordered, i.e., $P_{1, \theta}^{(e)} > ...> P_{K, \theta}^{(e)}$. Also, suppose we estimate the true class-wise error $P_{y, \theta}^{(e)}$ with $N$ i.i.d.~samples drawn from each class. Based on the estimated class-wise error $\hat{P}_{y, \theta}^{(e)}$, we pick the $M$ worst classes. If there are classes $y'$ and $y''$ with  $\hat{P}_{y', \theta}^{(e)}=\hat{P}_{y'', \theta}^{(e)}$, we randomly break the tie. Then, the probability that the true worst class belongs to the estimated $M$ worst classes, i.e., the probability of finding the worst class correctly is
\begin{align}
\Pr[ \text{ finding the worst class correctly } ] &= \sum_{m=1}^{M} \hat{P}_{1, \theta}^{(e),m} \\
&\ge \sum_{m=1}^{M} \eqref{eq:estimated as m-th worst}, \label{eq: probability of correct order}
\end{align}
where $\hat{P}_{1, \theta}^{(e),m}$ is the probability that $\hat{P}_{1, \theta}^{(e)}$ is the $m$-th largest value among $\hat{P}_{y, \theta}^{(e)}$, and \eqref{eq:estimated as m-th worst} is provided at the bottom of the page.
\end{theorem}
Theorem~\ref{thm:find the correct worst class} is a lower bound on the probability of finding the worst class correctly, stating two important points: 1) As the error probability of the worst class is typically much higher than the other classes, the probability of finding the worst class correctly is high since $\prod_{y' \ne 1,y_1,...,y_{m-1}} \Pr\left[\hat{P}_{1, \theta}^{(e)} > \hat{P}_{y', \theta}^{(e)}\right]$ in~\eqref{eq:estimated as m-th worst} is large. Note that $\prod_{y' \ne 1,y_1,...,y_{m-1}} \Pr\left[\hat{P}_{1, \theta}^{(e)} > \hat{P}_{y', \theta}^{(e)}\right]$ has more effect on~\eqref{eq:estimated as m-th worst} than $\prod_{y' \in \{y_1,...,y_{m-1}\}} \Pr\left[\hat{P}_{1, \theta}^{(e)} \le \hat{P}_{y', \theta}^{(e)}\right]$ since $m$ is smaller than the total number of class $K$. 2) As $M$ becomes larger, the probability of finding the correct worst class becomes higher since \eqref{eq: probability of correct order} is the sum over $m=1,\ldots,M$. Note that error probabilities of a few $M$ worst classes are much higher than the other classes when the model is trained under an imbalanced dataset with a small number of minority samples. For example, consider the error probabilities of the vanilla CE-trained model on CIFAR10 dataset in Table~\ref{tab:CIFAR10_0.01_step}. Then, the classification error probability $P_{y, \theta}^{(e)}$ for each $y$ is given by
\begin{align}
    \label{CIFAR10 class-wise error}
    [0.75, 0.67, 0.86, 0.96, 0.89, 0.06, 0.03, 0.05, 0.02, 0.03].
\end{align}
In this example, the worst class' error probability $0.96$ is much larger than the error probabilities of others. For more examples, see Section \ref{sec:experiment}, where the models trained with even SOTA loss functions show the same behavior. Thus, we can conclude that the failure probability of finding the worst class is very low even with a small number of minority samples.

\begin{figure}[t]
    \centering
    \includegraphics[width=0.9\linewidth]{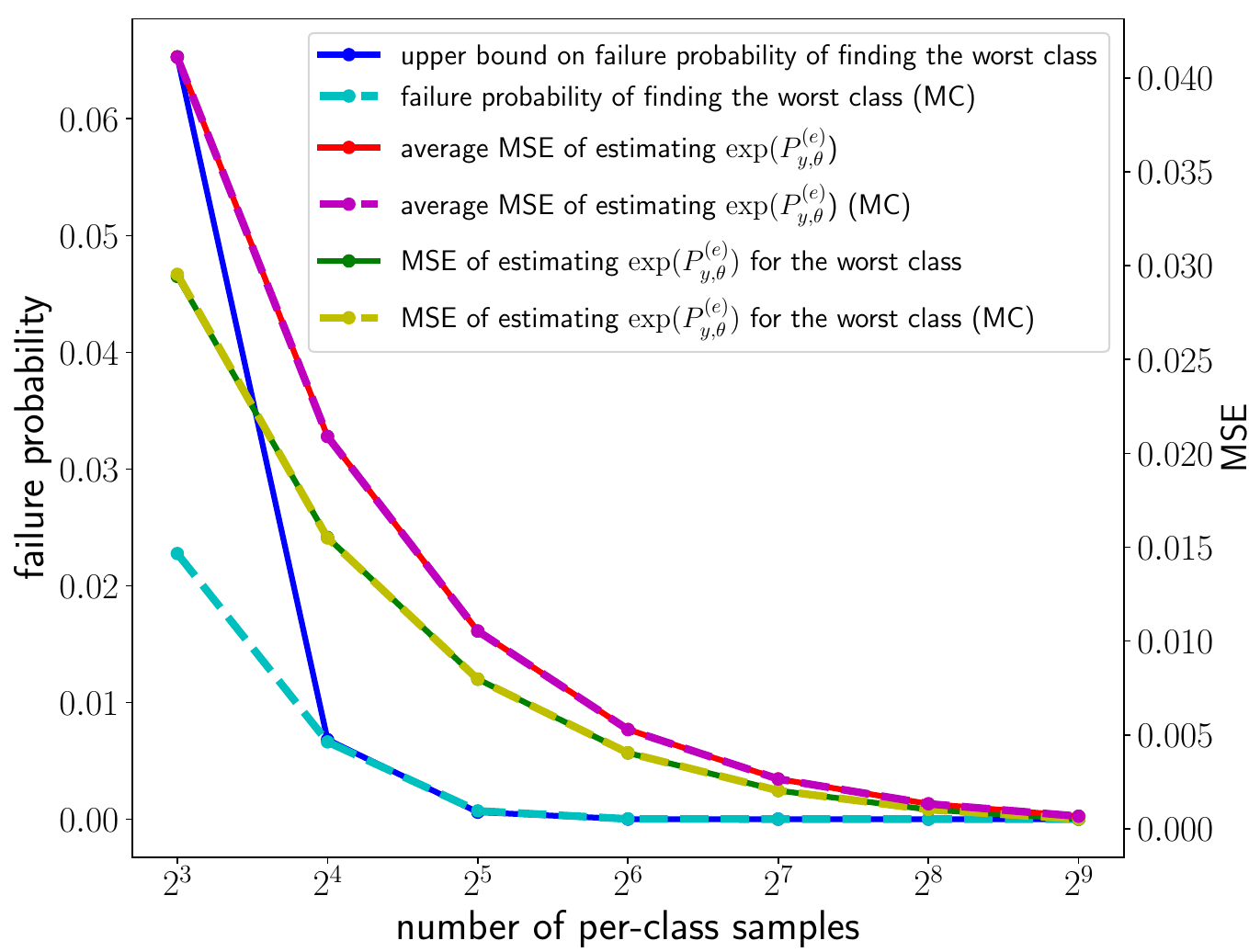}
    \caption{The calculation results of Theorems~\ref{thm:find the correct worst class} and~\ref{thm:MSE of error estimate} as well as the results of $10^6$ Monte Carlo experiments. The class-wise error probability is~\eqref{CIFAR10 class-wise error} and $M=3$. The solid lines are theoretical results, and the dotted lines are the results of the Monte Carlo experiments. When the number of samples is either far too few or far too many, all performances are poor or good, respectively. However, with a moderate number of samples, the failure probability remains relatively robust to the decrease in the number of samples. }\label{fig:thm_mc_val}
    \vspace{-0.15in}
\end{figure}

Next, we derive the MSE of $\exp{( \hat{P}_{y, \theta}^{(e)} )}$ with respect to the number of samples $N$ in the next theorem. Detailed proof is provided in Appendix~\ref{app:MSE of error estimate}.
\begin{theorem}
\label{thm:MSE of error estimate}
Suppose we estimate the true class-wise error probability $P_{y, \theta}^{(e)}$ with $N$ i.i.d.~samples drawn from class $y$. Then, the MSE of $\exp{( \hat{P}_{y, \theta}^{(e)} )}$ is
\begin{align}
&\sum_{n=1}^{N} \Pr \left[ \hat{P}_{y, \theta}^{(e)} = \frac{n}{N}    \right] \left( \exp{\left( P_{y, \theta}^{(e)} \right)} - \exp{\left( \frac{n}{N} \right)}   \right)^2\\
&=\sum_{n=1}^{N} \Bin(n,N,P_{y, \theta}^{(e)}) \left( \exp{\left( P_{y, \theta}^{(e)} \right)} - \exp{\left( \frac{n}{N} \right)}   \right)^2.
\end{align}
\end{theorem}
From the above theorem, we can identify two challenges in accurately estimating $\exp{( P_{y, \theta}^{(e)} )}$ compared to simply identifying the worst class: 1) When $N$ is small, $\exp{\left( \frac{n}{N} \right)}$ may not be close enough to $\exp{\left( P_{y, \theta}^{(e)} \right)}$ for every possible value of $n$, and 2) as $N$ decreases, the MSE increases directly due to \eqref{MSEerror}, since the estimation of the true $\exp{\left( P_{y, \theta}^{(e)} \right)}$ relies on $\hat{P}_{y, \theta}^{(e)}$. These limitations imply that the EGA method may struggle to identify the true ascent direction due to the high MSE of $\exp{( P_{y, \theta}^{(e)} )}$ especially when the sample size is small.

We verify the claim by theoretical evaluation and $10^6$ Monte Carlo experiments both. Figure~\ref{fig:thm_mc_val} shows the calculation results of Theorems~\ref{thm:find the correct worst class} and~\ref{thm:MSE of error estimate} as well as the results of $10^6$ Monte Carlo experiments. Unlike the failure probability of finding the worst class, which is almost $0$ when the number of per-class samples is larger than $16$, MSE of $\exp{\left( P_{y, \theta}^{(e)} \right)}$ grows exponentially as the number of per-class samples decreases. When the number of per-class samples is 16, the failure probability of finding the worst class is less than $0.01$, very low. On the contrary, MSE of $\exp{\left( P_{y, \theta}^{(e)} \right)}$ grows much faster. Note that obtaining $16$ or larger number of samples for minor classes is not a difficult task in practice. 

In summary, our linear ascent method is more robust than the EGA method with respect to the number of samples. This is because, when training on an imbalanced dataset with a small number of minority samples, our linear ascent can reliably find the true desired direction. This is possible by taking advantage of the fact that a few $M$ worst class errors are much larger than those of other classes, enabling us to identify the worst class with high probability. In contrast, the EGA method cannot utilize this fact to accurately estimate $\exp{\left( P_{y, \theta}^{(e)} \right)}$.

Thus, our linear ascent method increases the importance of the worst class more effectively than the EGA method especially when the number of samples is small. This successful increase in the importance of the worst class leads to higher worst-class accuracy in the min-max training. We will further verify this in Section~\ref{sec:minmax_result} by experiments.

\medskip
\noindent \textbf{Convergence Analysis:} Throughout the analysis, we assume that $R(\pi) := \min_{\theta} R(\pi,\theta)$ is differentiable with respect to $\pi$. As differentiability implies continuity, one can see that for any $0 < \epsilon < 1$, there exists $\pi^t$ such that 
\begin{align}
    \left\| \min_{\theta} R(\pi^*,\theta)-\min_{\theta} R(\pi^t,\theta) \right\| \le 2 \sqrt{2} K \sqrt{K} \epsilon,
\end{align}
where $\pi^*= \arg\max_{\pi} R(\pi) = \arg\max_\pi \min_{\theta} R(\pi,\theta)$ is the global optimal point. The next results show that the linear ascent algorithm finds such an optimal point within a certain accuracy. The proofs are in Appendices~\ref{app:linear ascent convergence area}~and~\ref{app:linear ascent convergence area2}, respectively.

\begin{theorem}
\label{thm:linear ascent convergence area}
Let $M=1$ and $\pi^{1}$ be an arbitrary initial prior in the probability simplex. Then, when $\alpha = \epsilon$ with $0<\epsilon<1$, there exists $t_0$ such that the linear ascent method satisfies 
\begin{align}
    \left\|\min_{\theta} R(\pi^*,\theta)-\min_{\theta} R(\pi^t,\theta) \right\| \le 2 \sqrt{2} K \sqrt{K} \epsilon
\end{align} for all $t>t_0$.
\end{theorem}

\begin{corollary}
\label{thm:linear ascent convergence area2}
Consider any $M \le K$, and let $\pi^{1}$ be an arbitrary initial prior in the restricted probability simplex such that $0 \le \pi^{1}_y \le \frac{1}{M}$ for all $y \in [K]$. Also, suppose that $\pi^*$ is also in the restricted simplex. Then, when $\alpha = \epsilon$ with $0<\epsilon<1$, there exists $t_0$ such that the linear ascent method satisfies 
\begin{align}
    \left\| \min_{\theta} R(\pi^*,\theta)-\min_{\theta} R(\pi^t,\theta) \right\| \le 2 \sqrt{2}\binom{K}{M} \sqrt{K} \epsilon
\end{align}
for all $t > t_0$.
\end{corollary}

A typical trajectory of $\pi^t$ is that it initially approaches $\pi^*$ and then stays around $\pi^*$. As the step size is $\alpha$, the trajectory forms at worst a regular $\binom{K}{M}$-side polygon including $\pi^*$ where each side has $2 \sqrt{2} \alpha$ length. Since $\pi^t$ stays close to $\pi^*$ and $R(\cdot)$ is Lipschitz continuous, the results hold. 

The bound obtained in Corollary~\ref{thm:linear ascent convergence area2} encourages choosing $M$ as small as possible, since the bound increases with $M$. Hence, by choosing $M=1$, one can guarantee the best performance in the sense of minimax accuracy as in Theorem~\ref{thm:linear ascent convergence area}. However, since $M=1$ only focuses on improving the accuracy of the single worst class at each time, it may require a number of iterations to find the global optimum. Choosing a larger (but not too large) $M$ could accelerate the speed of convergence as it boosts the accuracies of the $M$ worst classes simultaneously. The value of $M$ should be chosen so that it matches the size of the group with error probabilities similar to that of the worst class. This is because classes with similar error probabilities are likely to become the next worst class after an update. Therefore, to quickly find a good adversarial prior distribution, classes with similar performance to the worst class should be grouped and considered together. In the experiment, we also empirically observed that this method gives stable and good performance.

\subsection{Min-Max Algorithm Description}
\label{sec:alg}

Our min-max algorithm obtains $\max_{\pi} \min_{\theta} R(\pi,\theta)$ via iterating minimizing $R(\pi,\theta)$ with respect to the model parameter $\theta$ and maximizing $R(\pi,\theta)$ with respect to the target prior $\pi$. The minimizing step makes use of standard stochastic gradient descent (SGD) with the TLA loss function $l_{\TLA}(y,f(x) ; \pi^t)$ for given training data. The maximizing step is done by the linear ascent. Each step is performed on a different part of the training dataset.

\begin{algorithm}[t]
   \caption{Min-Max train algorithm}
\begin{algorithmic}
   \STATE {\bfseries Input:} $D_{\train}=\{(x_{n},y_{n})\}_{n=1}^{N}$, model parameter $\theta$, $T_0$, $T_1$, $T_2$, $M$, learning rate $\eta$, and $\alpha$

   \STATE {\bfseries Output:} the model satisfying Eq.~(\ref{eq:maxmin_formulation})

    \STATE Initialization: $\pi^{1} \leftarrow \pi^{\train}$, picks $\theta^0$ randomly
    \STATE Partition $D_{\train}$ into $D_{\model}$ and $D_{\prior}$

    \STATE \texttt{*warmup phase*}
    \FOR{$t=1$ {\bfseries to} $T_0$} 
    \STATE $g_{\theta}=\frac{1}{|D_{\model}|} \sum_{(x,y) \in D_{\model}} \nabla_{\theta} l_{\TLA}(y,f(x) ; \pi^t)|_{\theta=\theta^{t-1}}$
    \STATE $\theta^{t} \leftarrow \theta^{t-1}-\eta g_{\theta} $
    \STATE $\pi^{t+1} \leftarrow \pi^t$
    \ENDFOR
    
    \STATE \texttt{*minimax training*}
    \FOR{$t=T_0+1$ {\bfseries to} $T_0 + T_1$}
    \STATE $g_{\theta}=\frac{1}{|D_{\model}|} \sum_{(x,y) \in D_{\model}} \nabla_{\theta} l_{\TLA}(y,f(x) ; \pi^t)|_{\theta=\theta^{t-1}}$
    \STATE $\theta^{t} \leftarrow \theta^{t-1}-\eta g_{\theta} $
    \STATE Set $\pi^{t,\maxx,M}$ on $D_\prior$ by Eq.~(\ref{eq:linear_direction})
    \STATE $\pi^{t+1} \leftarrow \pi^t + \alpha (\pi^{t,\maxx,M} - \pi^t)$
    \ENDFOR
    
    \STATE \texttt{*fine tuning*}
    \FOR{$t=T_0 + T_1 + 1$ {\bfseries to} $T_0 + T_1 + T_2$}
    \STATE $g_{\theta}=\frac{1}{|D_{\train}|} \sum_{(x,y) \in D_{\train}} \nabla_{\theta} l_{\TLA}(y,f(x) ; \pi^t)|_{\theta=\theta^{t-1}}$
    \STATE $\theta^{t} \leftarrow \theta^{t-1}-\eta g_{\theta} $
    \STATE $\pi^{t+1} \leftarrow \pi^t$
    \ENDFOR
\end{algorithmic}
\label{alg:minmax}
\end{algorithm}

\medskip
\noindent \textbf{Partition Dataset:} 
As training proceeds, an overparameterized neural network obtains nearly $0\%$ training error~\cite{zhang2017understanding} while the test error stays away from zero. Hence, to measure the true $0$-$1$ loss reliably, one needs a new dataset that is unused in the training procedure. To this end, we partition the given training dataset $D_\train$ into two parts; $D_\model$ for model training and $D_\prior$ for finding an adversarial prior distribution. This technique is widely used in deep learning for updating hyperparameters of loss function~\cite{ren2018learning,liu2021improving,li2021autobalance}. To be specific, let $N_{y, \model}$ be the number of class-$y$ samples in $D_{\model}$ and $N_{y, \prior}$ be the number of class-$y$ samples in $D_{\prior}$. In this paper, inspired by the widely used ratio of train-validation data split, $0.8$ and $0.2$, we use $N_{y, \model}=0.8N_{y}$ and $N_{y, \prior}=0.2N_{y}$ where $N_{y}$ is the total number of training samples of class $y$. Note that it preserves the imbalance ratio after partitioning.

\medskip
\noindent \textbf{Overall Procedures:} The overall minimax algorithm consists of three phases. The first phase is the warmup phase of $T_0$ iterations, during which, the model is trained with a targeted loss function $l_{\TLA}(y,f(x) ; \pi^t)$. The target prior $\pi^t$ is fixed to be $\pi^\train$ for all $t \le T_0$. The second phase is the minimax training phase of the next $T_1$ iterations, i.e., from $t=T_0+1$ to $T_0 + T_1$, where the model is trained with $l_{\TLA}(y,f(x) ; \pi^t)$ and $\pi^{t}$ is updated by linear ascent for every iteration. This phase updates the target prior close to an adversarial prior, based on which trains the model. The last phase is the fine-tuning phase of $T_2$ iterations, where the model is fine-tuned on the full training data $D_{\train}$ without updating the target prior. This phase solves the problem of the small-size dataset resulted from data partition.

\section{Experimental Results}
\label{sec:experiment}

When a model is trained on imbalanced data with a vanilla training method, i.e., SGD with the CE loss, the worst-class accuracy is very low, see Tables~\ref{tab:CIFAR10}~and~\ref{tab:CIFAR100}. To solve this problem, we propose a min-max algorithm that uses the TLA loss function $l_{\TLA}(y,f(x) ; \pi^t)$ to minimize $0$-$1$ loss under the target prior $\pi^t$ in the minimizing step and the linear ascent to update the target prior $\pi^t$ close to the adversarial prior. 

In this experiment, we show that the widely used importance weighting method (TWCE) and the EGA method~\cite{zhang2020coping, sagawa2019distributionally, alaiz2005minimax, alaiz2007minimax} do not effectively maximize the worst-class accuracy when imbalanced small dataset is given. On the contrary, our proposed algorithm effectively maximizes the worst-class accuracy. In addition, we also show that current research for imbalanced data~\cite{cui2019class, cao2019learning, menon2020long, ye2020identifying, kini2021label} does not effectively increase the worst-class accuracy, especially in the case of a particular type of imbalance, the step imbalance. Our algorithm maximizes the worst-class accuracy while maintaining the class-balanced accuracy in a certain range.\footnote{ All codes are publicly available at https://github.com/hansung-choi/TLA-linear-ascent. }

\subsection{Experiment Setting}
\label{sec:Experiment Setting}
The experiment settings are similar to other imbalanced-data research for fair comparison~\cite{cui2019class, cao2019learning, menon2020long, ye2020identifying, kini2021label}. All experiments are repeated five times. 

\smallskip
\noindent \textbf{Datasets:} We evaluate our algorithm on CIFAR10 and CIFAR100 datasets~\cite{cifar10}. To introduce imbalanced data, we consider two types of imbalanced data: long-tail (LT) imbalance and step imbalance~\cite{cui2019class, cao2019learning, menon2020long, kini2021label}. In the LT-imbalanced data, the number of per-class samples exponentially decays from the head-to-tail classes. Specifically, $N_{y}=\rho^{\frac{y-1}{K-1}} N_{1}$ where $N_{1}$ is the same as in the original data and the imbalance ratio $\rho$ is set to $0.1, 0.01$ for CIFAR10 and $0.2, 0.1$ for CIFAR100, respectively. The step-imbalanced data are generated in a way that the first half of the classes, i.e., $1, 2, \ldots, \lceil K/2 \rceil$, is minor and has $N_\minor$ samples per class. The other half, classes $\lceil K/2 \rceil + 1, \ldots, K$, is major and has $N_\major$ samples per class. These major classes are the same as the original data. The imbalance ratio $\rho = \frac{N_\minor}{N_\major}$ is set to $0.1, 0.01$ for CIFAR10 and $0.2, 0.1$ for CIFAR100. Training data are randomly cropped and flipped horizontally to be consistent with previous research~\cite{cao2019learning, menon2020long, kini2021label}. Test data are left unmanipulated and used to estimate experiment results.

\smallskip
\noindent \textbf{Common Training Setting:} When we give no special description, the common training setting follows the next. We used ResNet32 \cite{he2016deep} as a model and SGD as an optimizer. The learning rate is $0.1$, momentum is $0.9$, weight decay is $2 \times 10^{-4}$, and the batch size is $128$. The learning rate is warmed up linearly for the first five epochs.

\smallskip
\noindent \textbf{Min-Max Training Setting:} To do an ablation study for verifying the advantage of our algorithm over the widely used importance weighting method (TWCE) and the EGA method~\cite{zhang2020coping, sagawa2019distributionally, alaiz2005minimax, alaiz2007minimax}, we replaced the TLA loss as the TWCE loss or the linear ascent as the EGA in Algorithm~\ref{alg:minmax}. In Algorithm~\ref{alg:minmax}, we set $T_0=5, T_1=295$, and $T_2=30$. The learning rate $\eta$ decayed by factor $0.01$ at the $200$-th and $320$-th epochs. The learning rates $\alpha$ of the linear ascent and the EGA are set as $0.01$ and $0.1$, respectively. The learning rate $\alpha$ is fixed as a constant value. The value $M$ of linear ascent is set as $3, 1$, and $10$ for LT-imbalance CIFAR10, step-imbalance CIFAR10, and CIFAR100, respectively. Hyperparameter $\tau$ of the TLA loss function is provided in Appendix~\ref{app:hyperparameters}.

\smallskip
\noindent \textbf{Approaches to Imbalanced Data:} Existing approaches to imbalanced data mainly focus on increasing the accuracy of the minor classes by modifying the CE loss. We train the model with these losses by SGD during the $300$ epoch, where the learning rate decayed by factor $0.01$ at the $160$-th and $220$-th epochs. Note that the generalized CE loss~\cite{kini2021label} can be written by
\begin{align}
    l(y,f(x)) = -w_{y}\log\left(\frac{\exp( \bigtriangleup_{y}f_{y}(x)+\ell_{y} )}{\sum_{y'}\exp( \bigtriangleup_{y'}f_{y'}(x)+\ell_{y'} )}\right).
\end{align}
The $9$ baseline loss functions are as follows.
\begin{enumerate}
    \item Vanilla CE: $w_{y}=\bigtriangleup_{y}=1$, $\ell_{y}=0$.

    \item Weighted CE (WCE): $w_{y}=1/N_y$, $\bigtriangleup_{y}=1$, $\ell_{y}=0$.

    \item Focal~\cite{lin2017focal}: $w_{y}=(1-\hat{p}_{y|x})^2$, $\bigtriangleup_{y}=1$, $\ell_{y}=0$ for all $y$ where $\hat{p}_{y|x}(y|x)=\frac{\exp( f_{y}(x) )}{\sum_{y'}\exp( f_{y'}(x) )}$.

    \item Focal-alpha: a combined version of WCE and focal losses, where $w_{y}=\frac{1}{N_{y}}(1-\hat{p}_{y|x})^{2}$, $\bigtriangleup_{y}=1$, $\ell_{y}=0$.

    \item LDAM~\cite{cao2019learning}: $w_{y}=\bigtriangleup_{y}=1$, $\ell_{y}=- C \cdot N_y^{-1/4}$ and $\ell_{y'}=0$ for $y' \ne y$, where $C$ is tuned so that $\max_{y}\left| \ell_{y} \right| = 0.5$. 

    \item LDAM-DRW~\cite{cao2019learning}: LDAM-DRW is a combined method of LDAM loss and a deferred re-weighting (DRW). DRW setting is the same as previous research \cite{cao2019learning}. setting $w_{y}=1$ of the LDAM loss during the first 160 epoch. Then setting $w_{y}=\frac{1-\beta}{1-\beta^{N_y}}$ of the LDAM loss until the experiment ends. In this experiment, $\beta=0.9999$. 

    \item LA~\cite{menon2020long}: $w_{y}=\bigtriangleup_{y}=1$, $\ell_{y}=\tau \log\pi^{\train}_{y}$. The value of $\tau$ is given in Appendix~\ref{app:hyperparameters}.

    \item VS~\cite{kini2021label}:  $w_{y}=1$, $\bigtriangleup_{y}=(N_{y}/N_{\max})^{\gamma}$, $\ell_{y}=\tau \log\pi^{\train}_{y}$. The values of $\gamma$ and $\tau$ are given in Appendix~\ref{app:hyperparameters}.
\end{enumerate}
The above loss functions are generalizations of the vanilla CE loss. A recent research proposed a new type of loss function that increases the accuracies of the worst class and the balanced class together~\cite{du2023no}, where geometric mean loss (GML) is newly proposed to maximize the geometric mean accuracy~\cite{du2023no}. Note that since the geometric mean is much more sensitive to the worst class than the arithmetic mean, GML effectively increases the worst-class accuracy. The GML is defined as follows.
\begin{enumerate}
\item[9)] GML~\cite{du2023no}: the GML for class $y$ is defined as
\begin{align}
l_{\GML}(y) = -\frac{1}{K}\sum_{y=1}^{K} \log \widehat{p}^{(c)}_y,
\end{align}
where $\widehat{p}^{(c)}_y$ is
\begin{align}
\widehat{p}^{(c)}_y =  \sum_{i=1}^{N_{b,y}} \frac{\exp(f_y(x^{i,y}))}{\sum_{y'}N_{b,y'}\exp(f_{y'}(x^{i,y}))}.
\end{align}
with $N_{b,y}$ being the number of samples of class $y$ in the mini-batch and $x^{i,y}$ is the $i$-th sample of class $y$ in the mini-batch. For LT-imbalance CIFAR10, setting the learning rate as $0.05$ proposed in the original paper~\cite{du2023no} gives better results than the common training setting. In other cases, setting the learning rate the same as the common training setting gives better results than setting the learning rate as $0.05$.
\end{enumerate}

We compared these approaches with our algorithm (TLA+linear ascent) and the widely used min-max method (TWCE+EGA~\cite{li2023wat, zhang2020coping, sagawa2019distributionally, alaiz2005minimax, alaiz2007minimax}), by which we demonstrate our proposed algorithm. 

\begin{table*}[ht]
\centering
\caption{The accuracy (\%) and target prior value (\%) of the worst class when different minimizing and maximizing methods are used in Algorithm~\ref{alg:minmax}. Our proposed method (TLA+linear ascent) shows the best worst-class accuracy for almost every case.}
\label{tab:minimax_worst_accuracy}
\begin{tabular}{|l|l|l|l|l|l|l|l|l|}
\hline
                     & \multicolumn{2}{c|}{TWCE+EGA~\cite{li2023wat, zhang2020coping, sagawa2019distributionally, alaiz2005minimax, alaiz2007minimax}}                 & \multicolumn{2}{c|}{TWCE+linear ascent}              & \multicolumn{2}{c|}{TLA+EGA}                           & \multicolumn{2}{c|}{TLA+linear ascent (ours)}                        \\ \hline
Metric               & Accuracy   & Prior value & Accuracy   & Prior value & Accuracy  & Prior value & Accuracy  & Prior value 
\\ \hline
CIFAR10-LT ($\rho=0.1$)    & $74.10${\tiny $\pm 2.22$} & $13.26${\tiny $\pm 12.32$} & $73.88${\tiny $\pm 1.58$} & $13.78${\tiny $\pm 11.74$} & $74.06${\tiny $\pm 1.16$} & $14.67${\tiny $\pm 9.78$}  & $\mathbf{74.48}${\tiny $\pm 1.55$} & $15.95${\tiny $\pm 10.05$} 
\\ \hline
CIFAR10-LT ($\rho=0.01$)   & $49.70${\tiny $\pm 7.15$} & $1.46${\tiny $\pm 1.80$}   & $54.38${\tiny $\pm 2.66$} & $2.60${\tiny $\pm 0.21$}   & $58.78${\tiny $\pm 2.04$} & $12.57${\tiny $\pm 3.92$}  & $\mathbf{59.78}${\tiny $\pm 3.28$} & $16.16${\tiny $\pm 3.11$}
\\ \hline
CIFAR10-step ($\rho=0.1$)  & $65.08${\tiny $\pm 2.82$} & $2.15${\tiny $\pm 2.62$}   & $66.96${\tiny $\pm 0.98$} & $17.25${\tiny $\pm 22.38$} & $\mathbf{70.68}${\tiny $\pm 1.64$} & $23.30${\tiny $\pm 8.80$}  & $70.22${\tiny $\pm 1.21$} & $19.03${\tiny $\pm 11.71$} 
\\ \hline
CIFAR10-step ($\rho=0.01$) & $27.40${\tiny $\pm 3.43$} & $44.80${\tiny $\pm 31.05$}  & $25.00${\tiny $\pm 3.47$} & $46.96${\tiny $\pm 21.92$} & $43.72${\tiny $\pm 3.91$}  & $10.82${\tiny $\pm 3.54$}  & $\mathbf{51.52}${\tiny $\pm 3.95$} & $18.44${\tiny $\pm 4.42$}  
\\ \hline
CIFAR100-LT ($\rho=0.2$)   & $18.60${\tiny $\pm 2.87$} & $0.11${\tiny $\pm 0.09$}   & $21.20${\tiny $\pm 2.14$} & $0.63${\tiny $\pm 0.42$}   & $19.40${\tiny $\pm 2.42$}   & $0.54${\tiny $\pm 0.48$}   & $\mathbf{24.00}${\tiny $\pm 1.90$} & $2.87${\tiny $\pm 2.71$}   
\\ \hline
CIFAR100-LT ($\rho=0.1$)   & $9.80${\tiny $\pm 1.60$}  & $0.09${\tiny $\pm 0.06$}   & $13.40${\tiny $\pm 2.58$} & $0.57${\tiny $\pm 0.46$}   & $12.80${\tiny $\pm 3.49$}   & $1.19${\tiny $\pm 1.21$}   & $\mathbf{19.20}${\tiny $\pm 1.83$} & $2.08${\tiny $\pm 1.04$}   
\\ \hline
CIFAR100-step ($\rho=0.2$) & $13.00${\tiny $\pm 6.72$} & $0.79${\tiny $\pm 1.09$}   & $17.60${\tiny $\pm 5.16$} & $0.55${\tiny $\pm 0.41$}   & $18.80${\tiny $\pm 2.99$}   & $1.20${\tiny $\pm 0.71$}   & $\mathbf{21.60}${\tiny $\pm 5.89$} & $2.41${\tiny $\pm 1.60$}    
\\ \hline
CIFAR100-step ($\rho=0.1$) & $0.20${\tiny $\pm 0.40$}  & $0.21${\tiny $\pm 0.12$}   & $14.20${\tiny $\pm 2.04$} & $0.25${\tiny $\pm 0.16$}   & $13.40${\tiny $\pm 4.22$}  & $0.43${\tiny $\pm 0.28$}   & $\mathbf{16.00}${\tiny $\pm 1.26$} & $1.29${\tiny $\pm 1.09$}   
\\ \hline
\end{tabular}
\vskip -0.1in
\end{table*}

\begin{table*}[ht]
\centering
\caption{The balanced accuracy (\%) when different minimizing and maximizing methods are used in Algorithm~\ref{alg:minmax}. Our proposed method (TLA+linear ascent) does not significantly degrade the balanced accuracy while improving the accuracy of the worst-performing class effectively (see Table~\ref{tab:minimax_worst_accuracy}).}
\label{tab:minimax_balanced_accuracy}
\begin{tabular}{|l|l|l|l|l|}
\hline
                     & TWCE+EGA~\cite{li2023wat, zhang2020coping, sagawa2019distributionally, alaiz2005minimax, alaiz2007minimax}   & TWCE+linear ascent         & TLA+EGA    & TLA+linear ascent (ours)          \\ \hline
CIFAR10-LT ($\rho=0.1$)    & $80.76${\tiny $\pm 0.53$} & $\mathbf{82.07}${\tiny $\pm 0.68$}          & $80.39${\tiny $\pm 0.60$} & $81.08${\tiny $\pm 0.54$} 
\\ \hline
CIFAR10-LT ($\rho=0.01$)   & $69.54${\tiny $\pm 1.09$} & $\mathbf{70.83}${\tiny $\pm 0.59$} & $69.39${\tiny $\pm 1.04$} & $70.78${\tiny $\pm 0.67$}          
\\ \hline
CIFAR10-step ($\rho=0.1$)  & $77.67${\tiny $\pm 1.16$} & $\mathbf{79.63}${\tiny $\pm 0.87$} & $78.70${\tiny $\pm 0.66$} & $78.88${\tiny $\pm 0.27$}          
\\ \hline
CIFAR10-step ($\rho=0.01$) & $64.47${\tiny $\pm 1.49$} & $63.11${\tiny $\pm 0.40$}          & $63.40${\tiny $\pm 0.91$}  & $\mathbf{65.91}${\tiny $\pm 1.48$} 
\\ \hline
CIFAR100-LT ($\rho=0.2$)   & $42.18${\tiny $\pm 0.79$} & $47.01${\tiny $\pm 0.56$}          & $45.29${\tiny $\pm 0.61$} & $\mathbf{48.36}${\tiny $\pm 0.41$} 
\\ \hline
CIFAR100-LT ($\rho=0.1$)   & $36.75${\tiny $\pm 0.79$} & $41.72${\tiny $\pm 0.49$}          & $39.48${\tiny $\pm 1.10$} & $\mathbf{43.09}${\tiny $\pm 0.77$} 
\\ \hline
CIFAR100-step ($\rho=0.2$) & $37.42${\tiny $\pm 5.06$} & $47.21${\tiny $\pm 0.82$}          & $46.05${\tiny $\pm 0.73$} & $\mathbf{51.15}${\tiny $\pm 0.43$} 
\\ \hline
CIFAR100-step ($\rho=0.1$) & $8.75${\tiny $\pm 8.40$}  & $43.49${\tiny $\pm 1.00$}          & $41.18${\tiny $\pm 0.75$} & $\mathbf{47.10}${\tiny $\pm 0.49$} 
\\ \hline
\end{tabular}
\vskip -0.1in
\end{table*}

\subsection{Results for Min-Max Approach}
\label{sec:minmax_result}
\begin{figure*}[ht]
\centering
\includegraphics[width=0.35\textwidth]{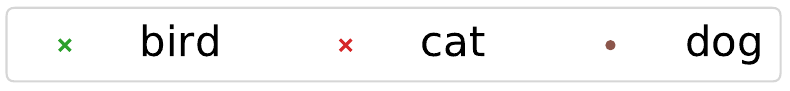}\\
\begin{multicols}{2}
\centering
\includegraphics[width=0.8\linewidth, height=0.30\textwidth]{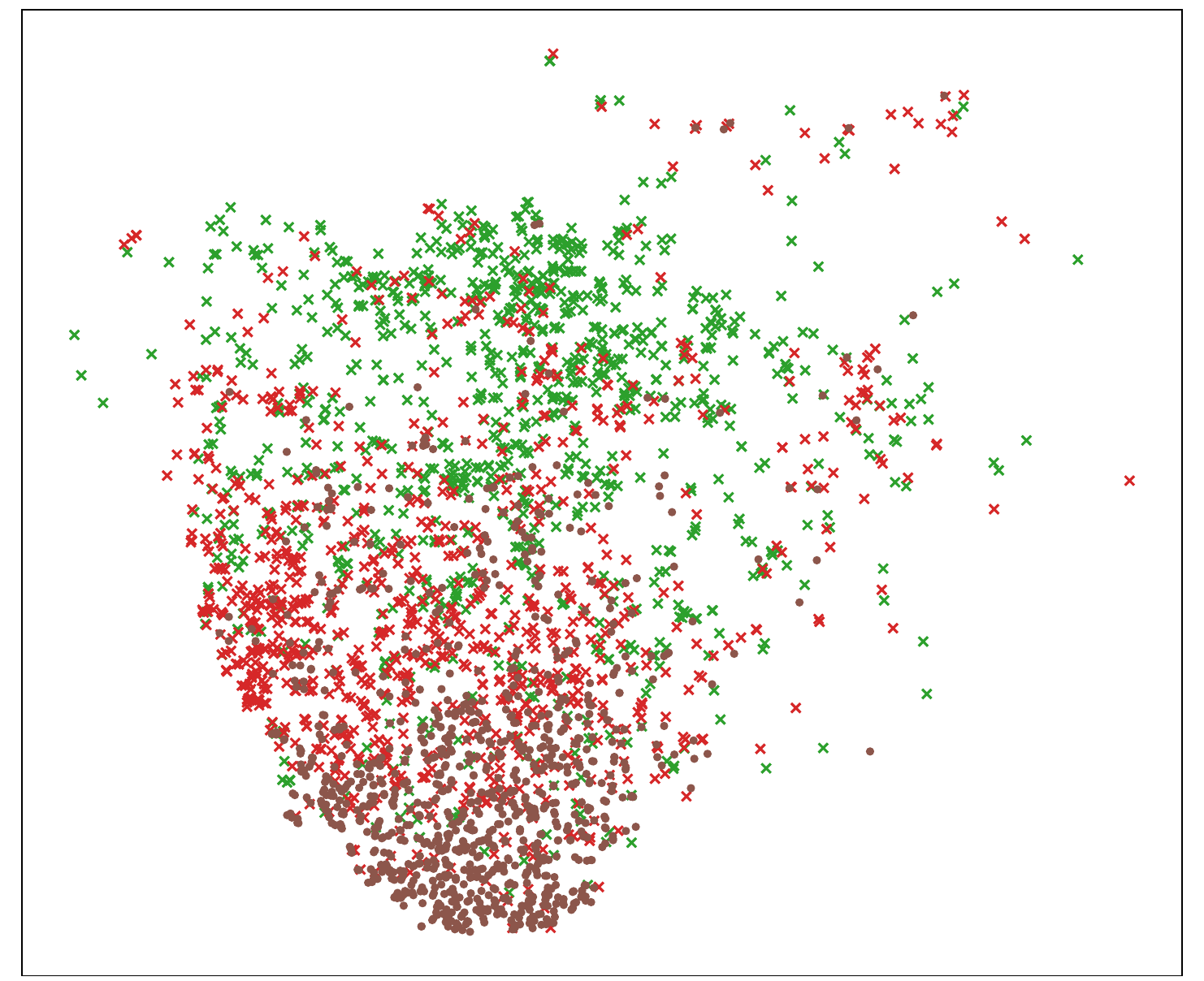}\\
Features from a model trained with the TLA loss

\includegraphics[width=0.8\linewidth, height=0.30\textwidth]{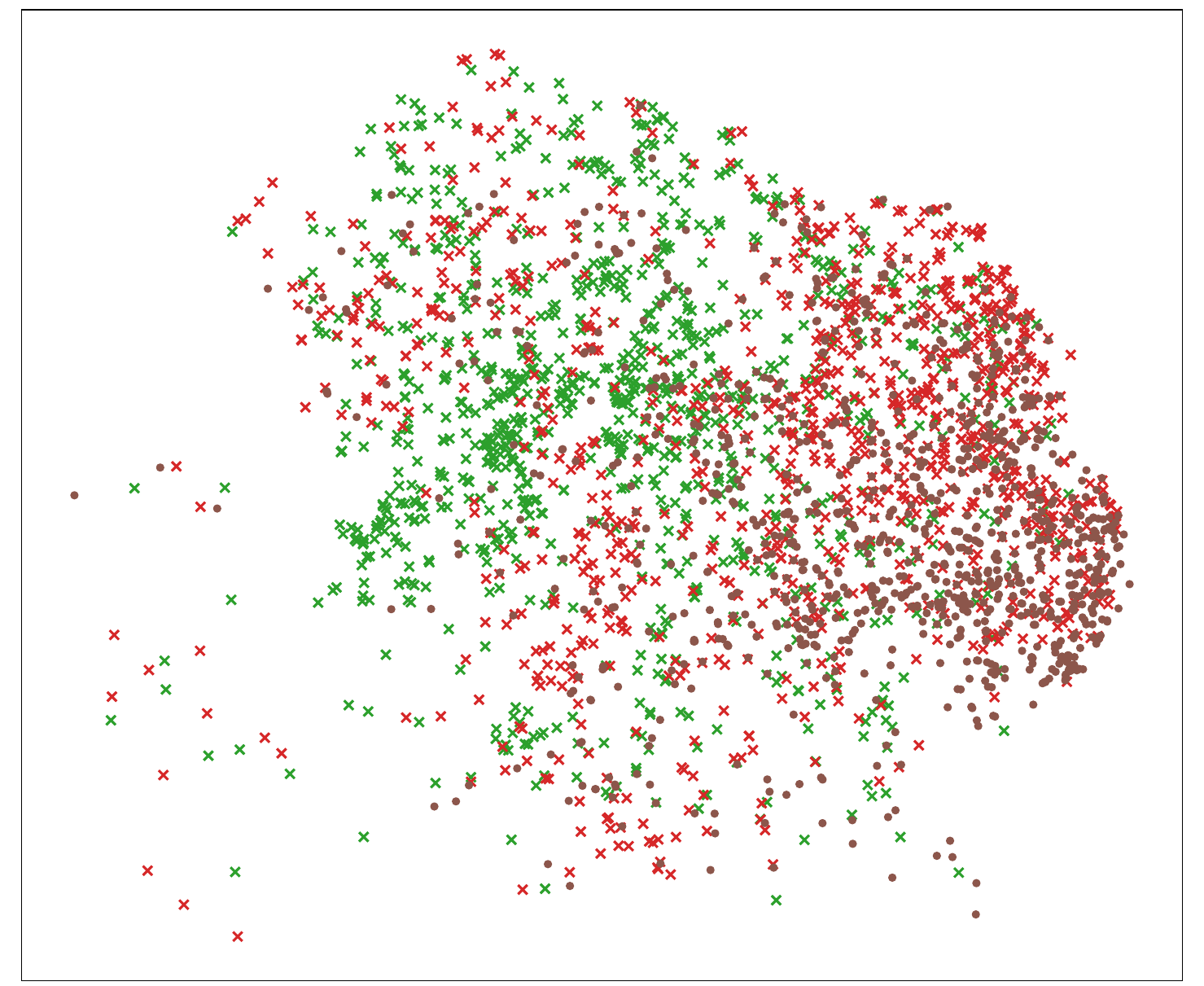}\\
Features from a model trained with the TWCE loss
\end{multicols}
\vspace{-0.15in}
\caption{Comparison via t-SNE between the TLA and TWCE losses for CIFAR10 step imbalance with $\rho=0.01$. In the case of TWCE, the features of cat class (red cross) widely overlap with those of dog class (brown dot). On the contrary, in the case of TLA, the features of cat class (red cross) are well clustered and have some distance from those of dog class (brown dot).}\label{fig:robust_loss_tsne}
\vspace{-0.15in}
\end{figure*}

Experimental results are summarized in Tables~\ref{tab:minimax_worst_accuracy} and~\ref{tab:minimax_balanced_accuracy}. Table~\ref{tab:minimax_worst_accuracy} shows the accuracy and the target prior value of the worst class, and Table~\ref{tab:minimax_balanced_accuracy} shows the class-balanced accuracy (balanced accuracy) when different loss functions and ascent methods are used in Algorithm~\ref{alg:minmax}. Our proposed method shows the highest or close to the highest worst-class accuracy and balanced accuracy. Our algorithm has advantages in the following perspectives.
\begin{itemize}
    \item Increasing the target prior value of the worst class: To increase the worst-class accuracy in the minimizing step, the target prior value of the worst class should be increased as large as possible in the maximizing step. As we verified in Figure~\ref{fig:thm_mc_val}, the linear ascent shows the largest prior value for the worst class except for CIFAR10-step dataset with $\rho=0.1$. The exceptional case ($\rho=0.1$) can be understood that the dataset has ``sufficient'' number of samples to estimate accuracy.

    \item Finding optimal decision boundaries: The model trained with the TLA loss function behaves closer to the optimal decision maker than the model trained with the TWCE loss function under the target prior.
    
    \item Produces better feature: The model trained with the TLA loss function can extract better features than the model trained with the TWCE loss function. This is investigated via t-SNE method in Figure~\ref{fig:robust_loss_tsne}.
\end{itemize}

\begin{table*}[ht]
\centering
\caption{Per-class accuracy (\%) of the experiment in Figure~\ref{fig:robust_loss_tsne}. The TLA loss increases the accuracy of bird class from 29.4\% to 63.8\% and the accuracy of cat class from 22.4\% to 65\% compared to those from the TWCE loss.}
\label{tab:cifar10_class_acc}

\begin{center}

\begin{tabular}{|l|l|l|l|l|l|l|l|l|l|l|}
\hline
      & plane & car  & bird & cat  & deer & dog  & frog & horse & ship & truck \\ \hline
TWCE & 43.4  & 59.5 & 29.4 & 22.4 & 35.9 & 74.6 & 73.4 & 79.6  & 73.1 & 90.4  \\ \hline
TLA   & 62.0  & 79.9 & 63.8 & 65.0 & 50.7 & 59.4 & 64.6 & 71.3  & 75.5 & 75.0    \\ \hline
\end{tabular}
\end{center}
\vspace{-0.15in}
\end{table*}

\smallskip
\noindent \textbf{Analysis:} The Bayes-optimal classification for the target prior $\pi^t$ can be made by choosing a class with the highest posterior distribution $p^t(y|x)$ among $y \in [K]$, by which it minimizes the total $0$-$1$ loss $R(\pi^t, \theta) = \sum_{y=1}^{K} \pi_{y}^t P^{(e)}_{y,\theta}$. Recall that
\begin{align}
p^t(y|x) = \frac{p(x|y) \pi_{y}^t}{ \sum_{y'} p(x|y') \pi_{y'}^t } = \frac{p(x|y) \pi_{y}^t}{p^t(x)}.
\end{align}
Thus, if we increase the target prior value for class $y$, then the Bayes-optimal classifier chooses class $y$ more frequently, which leads to an increase in the accuracy of class $y$.

This gives two perspectives for maximizing the worst-class accuracy. The first one is that the maximization method should increase the target prior value of the worst class as large as possible for a given classifier. Secondly, the minimization method should train the model close to the Bayes-optimal classifier under a fixed prior. In other words, the classifier increases the class accuracy when the class prior value is high and decreases the class accuracy when the class prior value is low to minimize the total $0$-$1$ loss. To this end, our proposed algorithm includes two components, TLA and linear ascent for minimization and maximization steps, respectively. 

To demonstrate our algorithm, first see Table~\ref{tab:minimax_worst_accuracy}. When the same loss function is used, the linear ascent increases the target prior value of the worst class more effectively than the EGA. Thus, linear ascent helps the model to increase the worst-class accuracy by helping the model focus on the worst class more efficiently than the other classes. In this sense, the linear ascent is a better method for the maximization step than the EGA.

The model trained with the TLA loss behaves like the optimal decision maker. For every case, the model trained with the TLA loss has the larger worst-class accuracy as the target prior value of the worst class increases. This is like the behavior of the above Bayes-optimal classifier, i.e., increasing the class accuracy when the class prior value is increased. On the contrary, the model trained with the TWCE loss behaves far from the optimal classifier. For some cases (the $3$-th, $6$-th, and $9$-th rows of Tabel~\ref{tab:minimax_worst_accuracy}), the model trained with the TWCE loss has the lower worst-class accuracy even when the target prior value of the worst class increased, which implies that the TWCE loss is strictly suboptimal. In addition, the model trained with the TWCE loss shows a lower worst-class accuracy than the model trained with the TLA loss for almost every case when the same ascent method is used. Hence, the model trained with the TLA loss function behaves closer to the optimal decision maker than the model trained with the TWCE loss function under the target prior.

Lastly, we compare extracted features from the models trained with the TLA and TWCE losses while fixing the linear ascent for the maximization step in Algorithm~\ref{alg:minmax}. First, see Table~\ref{tab:cifar10_class_acc} for a comparison of the TLA and TWCE losses on the step-imbalance CIFAR10 dataset at $\rho=0.01$. As shown, the model trained with the TLA loss increased the accuracy of the cat (worst) class from $22.4\%$ to $65.0\%$ and the bird (second worst) class from $29.4\%$ to $63.8\%$. This performance gain can be visually explained using the t-SNE method~\cite{van2008visualizing}, where features are taken from the input of the final layer. Figure~\ref{fig:robust_loss_tsne} depicts features in the t-SNE space, where only three classes (bird, cat, dog) are shown for visual brevity. The full t-SNE figure is available in Appendix~\ref{full_tsne_result}. For TWCE, a common classification error is between the cat and dog classes; the features of the cat (red) class significantly overlap with those of the dog (brown) class. In contrast, with TLA, the features of the cat (red) class are well clustered and maintain some distance from those of the dog (brown) class. The bird class is the second worst class, which locates across multiple classes.

To analyze this further, we numerically measured how well each class is clustered using the inter-intra distance method. Let $d_{\inter}(y)$ be the average Euclidean distance from the center of class $y$ to the centers of the three nearest classes. Similarly, let $d_{\intra}(y)$ be the average Euclidean distance from samples of class $y$ to their center. Our final metric is the ratio $r(y)=\frac{d_{\inter}(y)}{d_{\intra}(y)}$; a larger $r(y)$ implies a more clearly distinguishable feature cluster. The TLA loss increases $r(y)$ for the cat (worst) class by $58.94\%$, for the bird (2nd worst) class by $14.95\%$, for the deer (3rd worst) class by $74.23\%$, and for the dog (most indistinguishable from cat) class by $104.80\%$. The inter-intra distance analysis quantitatively confirms our visual findings in Figure~\ref{fig:robust_loss_tsne}, and hence, we can conclude that our TLA loss improves feature extraction as well.

\begin{table*}[ht]
\centering
\caption{The worst-class and balanced accuracies (\%) on the CIFAR10 dataset. Our algorithm (TLA) shows the best performance or is comparable to other baselines in the worst-class accuracy metric.}
\label{tab:CIFAR10}
\begin{tabular}{|l|l|l|l|l|l|l|l|l|}
\hline
Imbalance type    & \multicolumn{4}{c|}{CIFAR10-LT}  & \multicolumn{4}{c|}{CIFAR10-Step}                                                                       \\ \hline
Imbalance ratio   & \multicolumn{2}{c|}{0.1}  & \multicolumn{2}{c|}{0.01}  & \multicolumn{2}{c|}{0.1} & \multicolumn{2}{c|}{0.01}              \\ \hline
Metric            & Worst class    & Balanced       & Worst class    & Balanced  & Worst class & Balanced & Worst class & Balanced
\\ \hline
CE                & $72.46${\tiny $\pm 1.78$}   & $82.31${\tiny $\pm 0.19$}     & $35.58${\tiny $\pm 1.71$}       & $66.14${\tiny $\pm 0.39$}    & $33.18${\tiny $\pm 1.44$}    & $77.47${\tiny $\pm 0.44$}  & $3.16${\tiny $\pm 1.00$}    & $56.83${\tiny $\pm 0.55$} 
\\ \hline
WCE               & $72.16${\tiny $\pm 1.11$}       & $82.64${\tiny $\pm 0.39$}       & $40.38${\tiny $\pm 2.96$}        & $66.47${\tiny $\pm 1.00$}     & $37.52${\tiny $\pm 0.79$}       & $78.74${\tiny $\pm 0.18$}    & $6.14${\tiny $\pm 1.29$}   & $58.20${\tiny $\pm 0.99$}  
\\ \hline
Focal~\cite{lin2017focal}    & $69.70${\tiny $\pm 1.73$}         & $80.62${\tiny $\pm 0.78$}     & $31.98${\tiny $\pm 2.57$}        & $63.12${\tiny $\pm 0.78$}    & $30.88${\tiny $\pm 1.85$}       & $75.92${\tiny $\pm 0.33$}    & $2.08${\tiny $\pm 0.28$}    & $54.72${\tiny $\pm 0.28$} 
\\ \hline
Focal alpha~\cite{lin2017focal} & $70.26${\tiny $\pm 0.61$}         & $81.33${\tiny $\pm 0.29$}    & $38.86${\tiny $\pm 4.55$}         & $64.09${\tiny $\pm 0.75$}    & $37.40${\tiny $\pm 1.66$}       & $77.78${\tiny $\pm 0.47$}     & $4.04${\tiny $\pm 1.35$}  & $54.10${\tiny $\pm 0.77$} 
\\ \hline
LDAM~\cite{cao2019learning}  & $75.68${\tiny $\pm 1.20$}        & $84.93${\tiny $\pm 0.13$}    & $45.78${\tiny $\pm 2.45$}        & $71.09${\tiny $\pm 0.59$}     & $41.62${\tiny $\pm 0.93$}      & $80.61${\tiny $\pm 0.34$}     & $5.26${\tiny $\pm 0.84$}   & $59.02${\tiny $\pm 0.50$} 
\\ \hline
LDAM+DRW~\cite{cao2019learning} & $\mathbf{76.92}${\tiny $\pm 0.90$} & $\mathbf{85.98}${\tiny $\pm 0.26$} & $52.94${\tiny $\pm 3.80$}  & $73.70${\tiny $\pm 0.96$}    & $51.54${\tiny $\pm 1.17$}       & $82.33${\tiny $\pm 0.19$}     & $8.32${\tiny $\pm 0.75$}  & $60.92${\tiny $\pm 0.23$} 
\\ \hline
LA~\cite{menon2020long}  & $72.62${\tiny $\pm 0.75$}      & $84.34${\tiny $\pm 0.16$}    & $56.02${\tiny $\pm 4.04$}       & $74.26${\tiny $\pm 0.74$}   & $55.92${\tiny $\pm 2.28$}       & $\mathbf{82.66}${\tiny $\pm 0.18$} & $41.72${\tiny $\pm 3.52$}  & $\mathbf{72.22}${\tiny $\pm 0.27$}
\\ \hline
VS~\cite{kini2021label}  & $71.62${\tiny $\pm 1.23$}    & $84.44${\tiny $\pm 0.13$}     & $58.98${\tiny $\pm 3.29$}        & $\mathbf{74.80}${\tiny $\pm 0.65$} & $62.86${\tiny $\pm 1.14$}  & $82.39${\tiny $\pm 0.23$}   & $33.62${\tiny $\pm 6.81$}      & $70.63${\tiny $\pm 0.74$}  
\\ \hline
GML~\cite{du2023no} & $61.40${\tiny $\pm 1.69$}    & $76.95${\tiny $\pm 0.57$}     & $45.70${\tiny $\pm 5.53$}        & $63.85${\tiny $\pm 0.68$} & $65.58${\tiny $\pm 1.52$}  & $78.99${\tiny $\pm 0.17$}   & $32.28${\tiny $\pm 4.25$}      & $65.76${\tiny $\pm 0.91$}  
\\ \hline
TWCE+EGA~\cite{li2023wat, zhang2020coping, sagawa2019distributionally, alaiz2005minimax, alaiz2007minimax}  & $74.10${\tiny $\pm 2.22$}     & $80.76${\tiny $\pm 0.53$}     & $49.70${\tiny $\pm 7.15$}         & $69.54${\tiny $\pm 1.09$}      & $65.08${\tiny $\pm 2.82$}         & $77.67${\tiny $\pm 1.16$}     & $27.40${\tiny $\pm 3.43$}    & $64.47${\tiny $\pm 1.49$}
\\ \hline
TLA+linear ascent (ours) & $74.48${\tiny $\pm 1.55$}    & $81.08${\tiny $\pm 0.54$}     & $\mathbf{59.78}${\tiny $\pm 3.28$} & $70.78${\tiny $\pm 0.67$}    & $\mathbf{70.22}${\tiny $\pm 1.21$} & $78.88${\tiny $\pm 0.27$}     & $\mathbf{51.52}${\tiny $\pm 3.95$} & $65.91${\tiny $\pm 1.48$}  
\\ \hline
\end{tabular}
\end{table*}

\begin{table*}[ht]
\centering
\caption{The worst-class and balanced accuracies (\%) on the CIFAR100 dataset. Our algorithm (TLA) shows the best performance in the worst-class accuracy metric.}
\label{tab:CIFAR100}

\begin{tabular}{|l|l|l|l|l|l|l|l|l|}
\hline
Imbalance type           & \multicolumn{4}{c|}{CIFAR100-LT}  & \multicolumn{4}{c|}{CIFAR100-Step}   
\\ \hline
Imbalance ratio          & \multicolumn{2}{c|}{0.2}  & \multicolumn{2}{c|}{0.1} & \multicolumn{2}{c|}{0.2} & \multicolumn{2}{c|}{0.1} 
\\ \hline
Metric                   & Worst class & Balanced & Worst class & Balanced & Worst class & Balanced & Worst class & Balanced 
\\ \hline
CE                       & $15.20${\tiny $\pm 1.47$}       & $55.14${\tiny $\pm 0.19$}     & $9.80${\tiny $\pm 1.17$}       & $49.89${\tiny $\pm 0.36$}  & $11.40${\tiny $\pm 0.80$}      & $54.02${\tiny $\pm 0.46$}   & $5.60${\tiny $\pm 1.20$}    & $48.60${\tiny $\pm 0.45$}  
\\ \hline
WCE                      & $12.00${\tiny $\pm 2.28$}       & $54.54${\tiny $\pm 0.41$}     & $6.60${\tiny $\pm 1.20$}       & $48.83${\tiny $\pm 0.44$}  & $11.80${\tiny $\pm 2.14$}      & $53.23${\tiny $\pm 0.41$}   & $4.40${\tiny $\pm 1.62$}    & $46.20${\tiny $\pm 0.17$}  
\\ \hline
Focal~\cite{lin2017focal}    & $13.40${\tiny $\pm 1.02$}       & $54.39${\tiny $\pm 0.36$}     & $7.40${\tiny $\pm 3.20$}       & $48.38${\tiny $\pm 0.71$}  & $12.60${\tiny $\pm 1.36$}      & $53.10${\tiny $\pm 0.80$}   & $5.60${\tiny $\pm 1.20$}     & $47.59${\tiny $\pm 0.23$}    
\\ \hline
Focal alpha~\cite{lin2017focal} & $11.80${\tiny $\pm 2.32$}       & $53.85${\tiny $\pm 0.55$}     & $7.00${\tiny $\pm 0.63$}       & $47.90${\tiny $\pm 0.40$}  & $9.40${\tiny $\pm 1.20$}       & $52.11${\tiny $\pm 0.18$}   & $4.00${\tiny $\pm 0.89$}     & $45.65${\tiny $\pm 0.36$}   
\\ \hline
LDAM~\cite{cao2019learning}  & $3.40${\tiny $\pm 1.36$}        & $57.46${\tiny $\pm 0.46$}     & $1.40${\tiny $\pm 1.02$}       & $52.18${\tiny $\pm 0.21$}  & $1.60${\tiny $\pm 1.02$}       & $54.90${\tiny $\pm 0.35$}   & $0.40${\tiny $\pm 0.49$}     & $49.05${\tiny $\pm 0.27$}  
\\ \hline
LDAM+DRW~\cite{cao2019learning}  & $9.20${\tiny $\pm 1.94$}        & $\mathbf{58.02}${\tiny $\pm 0.44$}  & $5.00${\tiny $\pm 1.67$}  & $\mathbf{53.76}${\tiny $\pm 0.46$}  & $8.20${\tiny $\pm 1.72$}   & $\mathbf{56.95}${\tiny $\pm 0.14$}   & $5.60${\tiny $\pm 0.49$} & $\mathbf{52.53}${\tiny $\pm 0.23$}  
\\ \hline
LA~\cite{menon2020long}  & $22.20${\tiny $\pm 2.32$}       & $55.70${\tiny $\pm 0.16$}     & $16.80${\tiny $\pm 1.47$}      & $50.83${\tiny $\pm 0.49$}  & $18.20${\tiny $\pm 2.93$}      & $55.88${\tiny $\pm 0.27$}   & $13.00${\tiny $\pm 2.10$}    & $51.89${\tiny $\pm 0.44$}  
\\ \hline
VS~\cite{kini2021label}   & $19.20${\tiny $\pm 3.06$}       & $55.70${\tiny $\pm 0.22$}     & $16.40${\tiny $\pm 2.06$}      & $51.02${\tiny $\pm 0.52$}  & $20.00${\tiny $\pm 1.79$}      & $56.25${\tiny $\pm 0.36$}   & $13.60${\tiny $\pm 1.85$}    & $51.55${\tiny $\pm 0.42$}  
\\ \hline
GML~\cite{du2023no} & $22.00${\tiny $\pm 2.61$}    & $56.08${\tiny $\pm 0.63$}     & $16.00${\tiny $\pm 2.61$}        & $51.20${\tiny $\pm 0.50$} & $21.20${\tiny $\pm 1.17$}  & $55.79${\tiny $\pm 0.36$}   & $14.80${\tiny $\pm 2.32$}      & $51.75${\tiny $\pm 0.55$}  
\\ \hline
TWCE+EGA~\cite{li2023wat, zhang2020coping, sagawa2019distributionally, alaiz2005minimax, alaiz2007minimax}  & $18.60${\tiny $\pm 2.87$}       & $42.18${\tiny $\pm 0.79$}     & $9.80${\tiny $\pm 1.60$}       & $36.75${\tiny $\pm 0.79$}  & $13.00${\tiny $\pm 6.72$}      & $37.42${\tiny $\pm 5.06$}   & $0.20${\tiny $\pm 0.40$}     & $8.75${\tiny $\pm 8.40$}     
\\ \hline
TLA+linear ascent (ours) & $\mathbf{24.00}${\tiny $\pm 1.90$} & $48.36${\tiny $\pm 0.41$}  & $\mathbf{19.20}${\tiny $\pm 1.83$}   & $43.09${\tiny $\pm 0.77$}  & $\mathbf{21.60}${\tiny $\pm 5.89$}  & $51.15${\tiny $\pm 0.43$}   & $\mathbf{16.00}${\tiny $\pm 1.26$} & $47.10${\tiny $\pm 0.49$}  
\\ \hline
\end{tabular}
\vspace{-0.1in}
\end{table*}

\subsection{Results for Imbalanced-Data Approaches}
\label{sec:imbalanced_data_approach_result}
Experimental results are summarized in Tables~\ref{tab:CIFAR10}~and~\ref{tab:CIFAR100}. Table~\ref{tab:CIFAR10} shows the worst-class accuracy and the balanced accuracy results for the CIFAR10 dataset, and Table~\ref{tab:CIFAR100} shows the results for CIFAR100. Our proposed method always gets the best result in the worst-class accuracy metric except for the LT-imbalanced CIFAR10 case with imbalance ratio $0.1$, i.e., except for the case when the number of minority samples is sufficient.

First, note that the results of the widely used TWCE+EGA~\cite{li2023wat, zhang2020coping, sagawa2019distributionally, alaiz2005minimax, alaiz2007minimax} method are consistent with our analysis in Section~\ref{sec:related_work}. As the number of per-class samples decreases and data become more imbalanced, the TWCE loss damages feature quality more~\cite{ye2020identifying} and EGA converges not to the correct adversarial prior. In particular for the step-imbalanced CIFAR100 with $\rho = 0.1$, these side effects are more significant and make the model performance worse than that of the model with the CE loss. Tables~\ref{tab:CIFAR10} and~\ref{tab:CIFAR100} state that current imbalanced-data approaches~\cite{ cao2019learning, menon2020long, kini2021label} may have a limited effect on boosting the worst-class accuracy, compared to ours. In addition, widely used minimax approach (TWCE+EGA)~\cite{li2023wat, zhang2020coping, sagawa2019distributionally, alaiz2005minimax, alaiz2007minimax} cannot maximize the worst-class accuracy well when the number of per-class samples is small. On the contrary, our proposed algorithm (TLA+linear ascent) successfully maximizes the worst-class accuracy.

\section{Conclusion}
\label{sec:conclusion}
In this work, we have developed a new minimax training algorithm that utilizes a new loss function, referred to as the TLA loss, and a new prior update method, referred to as the linear ascent method. The TLA loss addresses the issues of failure on finding optimal decision boundaries and damaged feature extraction from widely used importance weighting technique. We also show that the TLA loss has a higher generalization ability than the weighting method based on our newly derived prior-dependent generalization bound. The linear ascent method updates the prior values based on whether a class is $M$ worst or not, rather than its error probability. Both components exhibit provable convergence properties. The entire algorithm appropriately iterates the minimizing step with the TLA loss and the maximizing step with the linear ascent. Empirical evaluation demonstrates the best or comparable performance to other baselines in terms of the worst-class accuracy. Our method can be a good alternative to current imbalanced-data approaches when maximizing the worst-class accuracy is an important goal or the true prior distribution of the test dataset is unknown.

\appendices

\section{Proof of Theorem~\ref{thm:gen2}}
\label{app:gen2}
The entire proof structure consists of successively deriving upper bounds. Firstly, we derive the upper bound on $L^{l}_{\pi}(f)$ in terms of $L^{l}_{\pi^\train}(f)$. Secondly,  we derive the upper bound on $L^{l}_{\pi^\train}(f)$ in terms of $\hat{\mathfrak{R}}_N(\mathcal{G}^l)$. Lastly, we derive the upper bound on $\hat{\mathfrak{R}}_N(\mathcal{G}^l)$ in terms of hyperparameters of the loss function $l(y, f(x))$.

The relation between $L^{l}_{\pi}(f)$ and $L^{l}_{\pi^\train}(f)$ is immediate via the next procedure.
\begin{align}
L^{l}_{\pi}(f)&= \sum_{y=1}^{K} \pi_y L^{l}_{y}(f)
= \sum_{y=1}^{K} \frac{\pi_y}{\pi^\train_y} \pi^\train_y L^{l}_{y}(f)\\
\label{eq:prior_upper_bound}
& \le \frac{\pi_{\maxx}}{\pi^{\train}_{\min}} \sum_{y=1}^{K} \pi^\train_y L^{l}_{y}(f)
= \frac{\pi_{\maxx}}{\pi^{\train}_{\min}} L^{l}_{\pi^\train}(f)
\end{align}
where $ L^{l}_{y}(f) := \E_{x\sim p(x|y)}\left[ l(y,f(x)) \right]$. The upper bound of $L^{l}_{\pi^\train}(f)$ is given in the previous generalization bound works~\cite{wang2023unified,mohri2018foundations} as the next equation.
\begin{align}
\label{eq:emp_upper_bound}
L^{l}_{\pi^\train}(f) \le \hat{L}^{l}(f) + \hat{\mathfrak{R}}_N(\mathcal{G}^l) + 3B \sqrt{\frac{\log{2/\delta}}{2N}}.
\end{align}

Now we derive the upper bound on $\hat{\mathfrak{R}}_N(\mathcal{G}^l)$ based on the results from previous work~\cite{wang2023unified}.
In previous work, the upper bound on $\hat{\mathfrak{R}}_N(\mathcal{G}^l)$ is written as
\begin{align}
\label{eq:emp_Rad_upperbound}
\hat{\mathfrak{R}}_N(\mathcal{G}^l) \le \hat{\mathfrak{R}}_N(\mathcal{F}) \sum_{y=1}^K \mu_y \sqrt{\pi^{\train}_y}
\end{align}
where $\mu_y$ is a class-wise constant that depends on the data distribution and the loss function. Considering an arbitrary data sample $(x,y)$, $\mu_y = \mu_y(x)$ can be bounded as follows, e.g.,~\cite{wang2023unified}:
\begin{align}
\label{eq:sample_constant_upperbound}
\mu_y(x) \le 
w_y \overline{\Delta_y} \left[ 1-\softmax(\Delta_y f_y(x) + \ell_y)  \right].
\end{align}
Then, taking the supremum of the right side of~\eqref{eq:sample_constant_upperbound} for every possible data sample, we can find the upper bound on $\mu_y$.
\begin{align}
\label{eq:constant_upperbound}
\mu_y \le 
w_y \overline{\Delta_y} \left[ 1-\softmax(\Delta_y \min_{x \sim p(x|y)} f_y(x) + \ell_y)  \right].
\end{align}

Combining all bounds together, we complete the proof of Theorem~\ref{thm:gen2}.

\section{Proof of Theorem \ref{thm:find the correct worst class}}
\label{app:find the correct worst class}

Based on $N$ samples from each class $y$ and~\eqref{error estimate}, we estimate $P_{y, \theta}^{(e)}$, the true error probability of class $y$ at model parameter $\theta$. Then, we get $\left\{ \hat{P}_{1, \theta}^{(e)},...,\hat{P}_{K, \theta}^{(e)}  \right\}$. Based on this estimation, we reorder $\left\{ \hat{P}_{1, \theta}^{(e)},...,\hat{P}_{K, \theta}^{(e)}  \right\}$ in descending order as $\left\{ \hat{P}_{y_1, \theta}^{(e)},...,\hat{P}_{y_K, \theta}^{(e)}  \right\}$. If there are classes $y'$ and $y''$ with  $\hat{P}_{y', \theta}^{(e)}=\hat{P}_{y'', \theta}^{(e)}$, the order can be randomly determined. In addition to this, we assume the adversarial random scenario, where randomness tries to place the worst class $1$ as far back as possible, i.e, if $\hat{P}_{1, \theta}^{(e)}=\hat{P}_{y, \theta}^{(e)}$, adversarial randomness decides that the order of the worst class $1$ is behind of the class $y$. As the probability that the order of the worst class $1$ is later becomes larger, the probability that the true worst class belongs to the estimated $M$ worst classes becomes smaller. Thus, this assumption decreases the probability of finding the correct worst class.

Let $\hat{P}_{1, \theta}^{(e),m}$ be the probability such that $\hat{P}_{1, \theta}^{(e)}$ is the $m$-th largest value among $\hat{P}_{y, \theta}^{(e)}$. With this notation, the probability of finding the correct worst class can be written as follows.
\begin{align}
\Pr[ \text{finding the correct worst class} ] = \sum_{m=1}^{M} \hat{P}_{1, \theta}^{(e),m}
\end{align} 
For $\hat{P}_{1, \theta}^{(e)}$ to be estimated as the $m$-th largest value, $\hat{P}_{1, \theta}^{(e)}$ should be equal or smaller than $\hat{P}_{y_1, \theta}^{(e)},...,\hat{P}_{y_{m-1}, \theta}^{(e)}$ and $\hat{P}_{1, \theta}^{(e)}$ should be larger than $\hat{P}_{y_{m+1}, \theta}^{(e)},...,\hat{P}_{y_K, \theta}^{(e)}$ under our adversarial random scenario. Also, note that
\begin{align}
&\Pr \left[ \hat{P}_{1, \theta}^{(e)} > \hat{P}_{y_{m+1}, \theta}^{(e)}~\text{and}~\hat{P}_{1, \theta}^{(e)} > \hat{P}_{y_{m+2}, \theta}^{(e)} \right]\\
=& \Pr \left[ \hat{P}_{1, \theta}^{(e)} > \hat{P}_{y_{m+1}, \theta}^{(e)} \right] \Pr \left[ \hat{P}_{1, \theta}^{(e)} > \hat{P}_{y_{m+2}, \theta}^{(e)} \right]
\end{align} 
holds since $\hat{P}_{1, \theta}^{(e)},...,\hat{P}_{K, \theta}^{(e)}$ are independent of each other. Thus, $\hat{P}_{1, \theta}^{(e),m}$ under our adversarial random scenario which decreases the probability of finding the correct worst class can be described as~\eqref{eq:estimated as m-th worst}. Note that since the true $\hat{P}_{1, \theta}^{(e),m}$ is larger than $\hat{P}_{1, \theta}^{(e),m}$ derived from our adversarial random scenario,~\eqref{eq: probability of correct order} holds.

Now we calculate $\Pr\left[\hat{P}_{y, \theta}^{(e)} > \hat{P}_{y', \theta}^{(e)}\right]$. Recall that 
\begin{align}
    \label{app:error estimate}
    \widehat{P}^{(e)}_{y,\theta}=\frac{1}{N}\sum_{n=1}^{N}\mathds{1}\left[\widehat{y}(x_n;\theta) \ne y \right],
\end{align}
for $\{x_n\}_{n=1}^{N}$ sampled from class $y$. Since error occurs with probability $P_{y, \theta}^{(e)}$ for each sampled $x_n$, $\mathds{1}\left[\widehat{y}(x_n;\theta) \ne y \right]$ follows the Bernoulli distribution with probability $P_{y, \theta}^{(e)}$, i.e., $\Bern(P_{y, \theta}^{(e)})$. Then, $\sum_{n=1}^{N}\mathds{1}\left[\widehat{y}(x_n;\theta) \ne y \right]$ follows $\Bin\left(\cdot,N,P_{y, \theta}^{(e)}\right)$ since the sum of i.i.d. random variables of Bernoulli distribution follows Binomial distribution. Then, 
\begin{align}
&\Pr\left[\hat{P}_{y, \theta}^{(e)} > \hat{P}_{y', \theta}^{(e)}\right]\\
=&\Pr\left[\sum_{n=1}^{N} \mathds{1}\left[\widehat{y}(x_n;\theta) \ne y \right] > \sum_{n=1}^{N} \mathds{1}\left[\widehat{y'}(x'_n;\theta) \ne y' \right]\right]\\
=&\sum_{i=0}^{N-1} \sum_{n=1}^{N-i} \Bin\left( i+n,N,P_{y, \theta}^{(e)} \right) \Bin\left(i,N,P_{y', \theta}^{(e)} \right).
\end{align}
By applying the same procedures, we can also find $\Pr\left[\hat{P}_{y, \theta}^{(e)} \le \hat{P}_{y', \theta}^{(e)}\right]$, which completes the proof.

\section{Proof of Theorem~\ref{thm:MSE of error estimate}}
\label{app:MSE of error estimate}
We showed $\sum_{n=1}^{N} \mathds{1}\left[\widehat{y}(x_n;\theta) \ne y \right]$ of~\eqref{app:error estimate} follows $\Bin\left(\cdot,N,P_{y, \theta}^{(e)}\right)$ in Appendix~\ref{app:find the correct worst class}. Therefore,
\begin{align}
\Pr \left[ \hat{P}_{y, \theta}^{(e)} = \frac{n}{N}    \right] = \Pr \left[N \hat{P}_{y, \theta}^{(e)} = n    \right] = \Bin(n,N,P_{y, \theta}).
\end{align}
Using this, the MSE of $\exp{( \hat{P}_{y, \theta}^{(e)} )}$ can be derived directly as follows
\begin{align}
&\E\left[\left( \exp{\left( P_{y, \theta}^{(e)} \right)} - \exp{( \hat{P}_{y, \theta}^{(e)} )} \right)^2 \right] \\
&=\sum_{n=0}^{N} \Pr \left[ \hat{P}_{y, \theta}^{(e)} = \frac{n}{N}    \right] \left( \exp{\left( P_{y, \theta}^{(e)} \right)} - \exp{\left( \frac{n}{N} \right)}   \right)^2\\
&=\sum_{n=0}^{N} \Bin(n,N,P_{y, \theta}^{(e)}) \left( \exp{\left( P_{y, \theta}^{(e)} \right)} - \exp{\left( \frac{n}{N} \right)}   \right)^2.
\end{align}
It completes the proof.

\section{Proof of Theorem \ref{thm:linear ascent convergence area}}
\label{app:linear ascent convergence area}

To prove the claim, we first derive the upper bound on $\| \pi^* - \pi^{t} \|$ for $t > t_0$, where $\pi^*= \arg\max_{\pi}\min_{\theta} R(\pi,\theta)$ and $t_0$ is sufficiently large. Then, the Lipschitzness of $\min_\theta R(\pi, \theta)$ in turn proves the claim. Throughout the proof, we assume that $d$ is a feasible direction in the probability simplex, i.e., if $\pi^{t+1} = \pi^{t} + \alpha d$, then $\pi^{t+1}$ is also in the probability simplex. Note that the direction of linear ascent is always feasible.

We use the next two properties to derive the meaning of the direction in the linear ascent. The direction of $\pi^{t,\maxx,1} - \pi^t$ in the linear ascent is the direction in which $\pi^t$ approaches to $\pi^*$.

\begin{property}
\label{thm:ascent_R_prop}
Let $\theta^{t}=\arg\min_{\theta}R(\pi^{t},\theta)$. Then, a direction $d$ is an ascent direction of $\min_{\theta} R(\pi,\theta)$ at $\pi=\pi^{t}$ if $d$ is an ascent direction of $R(\pi,\theta^{t})$ at $\pi=\pi^{t}$. Likewise, a direction $d$ is an descent direction of $\min_{\theta} R(\pi,\theta)$ at $\pi=\pi^{t}$ if $d$ is an descent direction of $R(\pi,\theta^{t})$ at $\pi=\pi^{t}$.
\end{property}
\begin{IEEEproof}
    This property holds since $\nabla_{\pi}\min_{\theta}R(\pi,\theta)=\nabla_{\pi}R(\pi,\theta^{t})$ at $\pi=\pi^{t}$.
\end{IEEEproof}

\begin{property}
\label{thm:ascent_pi_prop}
Let $\pi^{t+1} = \pi^{t} + \alpha d$ for $\alpha > 0$. If $d$ is a direction that makes $\pi^{t+1}$ far from $\pi^*$ than $\pi^t$, then $d$ is an descent direction of $\min_{\theta} R(\pi,\theta)$ at $\pi=\pi^{t}$.
\end{property}
\begin{IEEEproof}
    This property holds since $\min_{\theta} R(\pi,\theta)$ is concave in $\pi$~\cite{gilet2020discrete}.
\end{IEEEproof}

\begin{figure}
    \centering
    \includegraphics[width=0.6\linewidth]{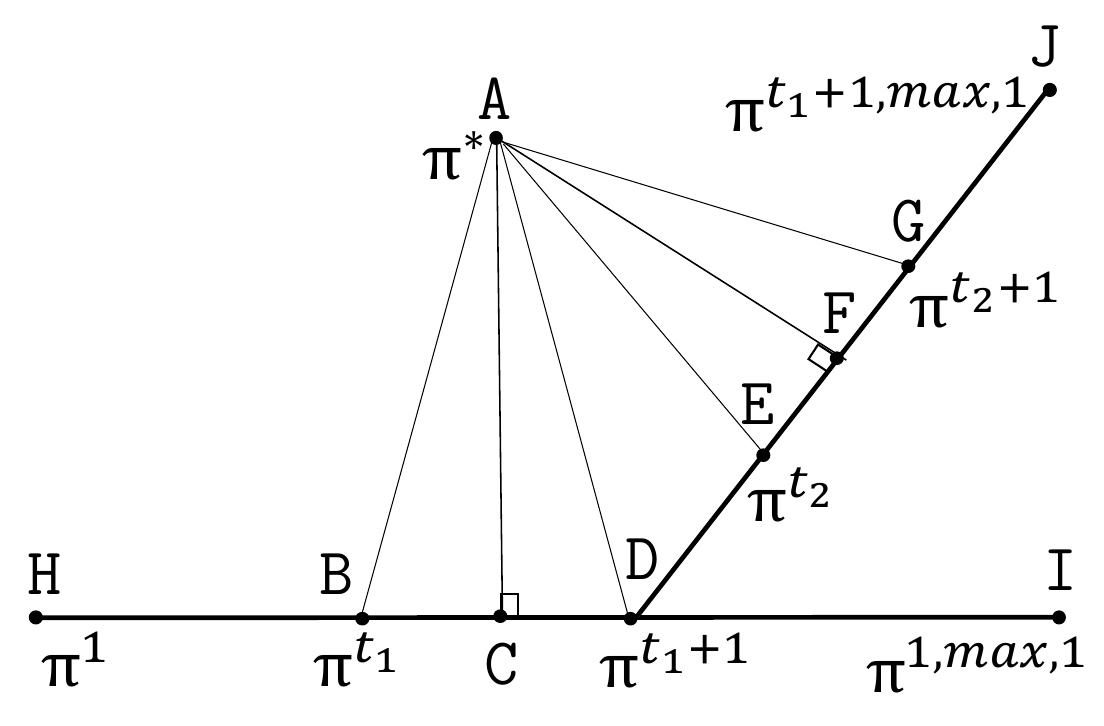}
    \caption{Adversarial scenario for convergence of linear ascent where $\pi^t$ moves as far away from $\pi^*$ as possible, subject to the constraints of linear ascent properties.}\label{fig:convergence_area_1}
    \vspace{-0.1in}
\end{figure}

\begin{figure}[t]
    \centering
    \includegraphics[width=0.3\linewidth]{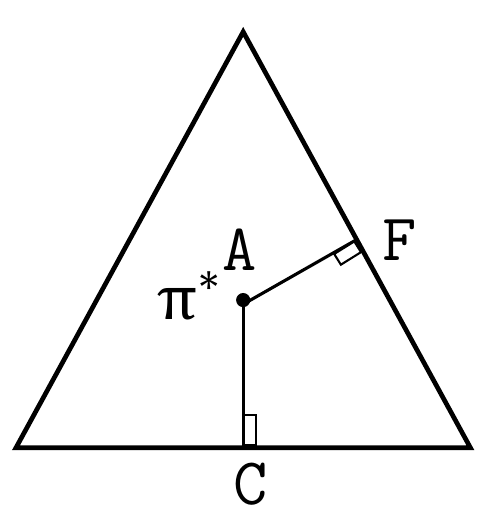}
    \caption{Adversarial scenario for convergence of linear ascent when $K=3$ after sufficiently many iterations.}\label{fig:convergence_area_2}
    \vspace{-0.1in}
\end{figure}

As $\pi^{t,\maxx,1}$ picks the worst-performing class, the direction of $\pi^{t,\maxx,1} - \pi^t$ is an ascent direction of $R(\pi,\theta^t)$ at $\pi=\pi^t$. Thus, the direction of $\pi^{t,\maxx,1} - \pi^t$ is an ascent direction of $\min_{\theta} R(\pi,\theta)$ at $\pi=\pi^{t}$ as well by Property~\ref{thm:ascent_R_prop}. This result together with  Property~\ref{thm:ascent_pi_prop} implies that the direction of $\pi^{t,\maxx,1} - \pi^t$ is always the direction in which $\pi^t$ approaches to $\pi^*$.

To derive the upper bound of $\| \pi^* - \pi^{t} \|$, it is sufficient to see the upper bound of the adversarial case. In the adversarial case, $\pi^{t}$ is updated as far as possible under the constraint that the direction of $\pi^{t,\maxx,1} - \pi^t$ is the direction in which $\pi^t$ approaches to $\pi^*$. Now, suppose that the linear ascent is iterated from an arbitrary prior $\pi^1$. Figure \ref{fig:convergence_area_1} illustrates the behavior of $\pi^t$ at an arbitrary $t$. The linear ascent shifts $\pi^t$ to $\pi^{1,\maxx,1}$ until $\pi^{t,\maxx,1}$ changes and monotonically decreases $\| \pi^* - \pi^{t} \|$ until $\pi^t$ passes point $C$, the closest point to $\pi^*$ on the line segment between $\pi^{t,\maxx,1}$ and $\pi^t$. If $\pi^{t,\maxx,1}$ is changed before $\pi^t$ passes the closest point to $\pi^*$ on the line segment between $\pi^{t,\maxx,1}$ and $\pi^t$, an adversarial situation does not occur. In this case, the distance $| \pi^* - \pi^{t} |$ decreases always. Therefore, for our adversarial scenario, $\pi^{t}$ should reach point $D$ instead.

To simplify notation, let point $B$ be $\pi^{t_1}$ and point $D$ be $\pi^{t_1+1}$, as marked in Figure~\ref{fig:convergence_area_1}. At point $D$, $\pi^{t_1+1,\maxx,1}$ changes from $\pi^{t_1,\maxx,1}$ since $\pi^{1,\maxx,1} - \pi^{t_1+1}$ is no longer an ascent direction of $\min R(\pi,\theta)$ at $\pi^{t_1+1}$ by Property~\ref{thm:ascent_pi_prop}. By repeating the above argument, we can get $\pi^{t_2+1}$ where $\pi^{t_2+1,\maxx,1}$ changes from $\pi^{t_1+1,\maxx,1}$. In the $i$-th argument, mark $\pi^{t_i+1}$ as the prior where $\pi^{t_i+1,\maxx,1}$ is different from $\pi^{t_i,\maxx,1}$.

Let point $E$ be $\pi^{t_2}$, point G be $\pi^{t_2+1}$, and point $F$ be the closest point to $\pi^*$ on the line segment between $\pi^{t_1+1,\maxx,1}$ and $\pi^{t_1+1}$. Since $\angle ABD \le \frac{\pi}{2} \le \angle AED$,
\begin{align}
    \overline{AB}^2+\overline{BD}^2 \ge \overline{AD}^2 \ge \overline{AE}^2+\overline{ED}^2.
\end{align}
By using  $\overline{AB}^2 + \overline{BD}^2 - \overline{ED}^2 \ge \overline{AE}^2$, 
\begin{align}
    &(\pi^*-\pi^{t_1})^2 - [(\pi^{t_2}-\pi^{t_1+1})^2-\alpha^2(\pi^{t_1,\maxx,1} - \pi^{t_1})^2]\\
    &\ge (\pi^*-\pi^{t_2})^2.
\end{align}
Thus, $(\pi^*-\pi^{t_2})^2$ is smaller than $(\pi^*-\pi^{t_1})^2$ if $(\pi^{t_2}-\pi^{t_1+1})^2 > \alpha^2(\pi^{t_1,\maxx,1} - \pi^{t_1})^2$. Likewise, $(\pi^*-\pi^{t_1})^2 > ... > (\pi^*-\pi^{t_{n-1}})^2$ when $(\pi^{t_{i+1}}-\pi^{t_i+1})^2 > \alpha^2(\pi^{t_i,\maxx,1} - \pi^{t_i})^2$ holds for $1 \le i \le n-1$. Assume that $(\pi^{t_{i+1}}-\pi^{t_i+1})^2 \le \alpha^2(\pi^{t_i,\maxx,1} - \pi^{t_i})^2$ at $i=n$. Then, $\|\pi^{t_{n+1}}-\pi^{t_n+1}\| \le \sqrt{2} \alpha$. This concludes,
\begin{align}
&\|\pi^{t_{n+1}+1}-\pi^{t_n+1}\| \\
&\le \|\pi^{t_{n+1}+1}-\pi^{t_{n+1}}\| + \|\pi^{t_{n+1}}-\pi^{t_n+1}\| \le 2 \sqrt{2} \alpha. 
\end{align}
If such $n$ does not exist, the distance $\|\pi^*-\pi^{t_i}\|$ always monotonically decreases for all $0 \le i$, i.e., this case is not an adversarial scenario. Now we can derive the adversarial case. The line segments of $\pi^{t_{i+1}+1}-\pi^{t_i+1}$ surround $\pi^*$ and cycles per $P$ repetitions by forming a regular $P$ sided polygon with side length $2 \sqrt{2} \alpha$ for all $n \le i$. If not, some $\pi^t$ can enter in this $P$ sided polygon area. In this case, $\pi^t$ approaches closer to $\pi^*$ than the adversarial case. 

\begin{figure*}[t]
\centering
\includegraphics[width=0.55\textwidth]{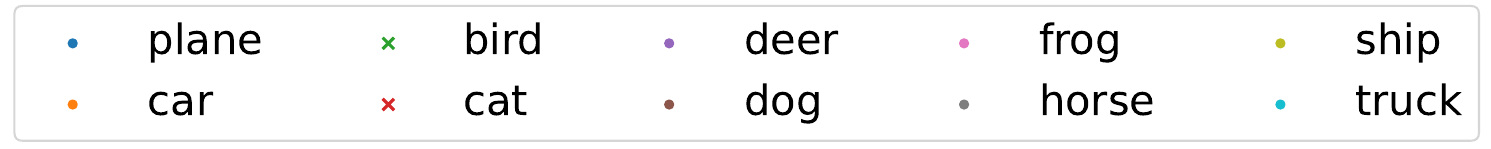}\\
\begin{multicols}{2}
\centering
\includegraphics[width=0.7\linewidth]{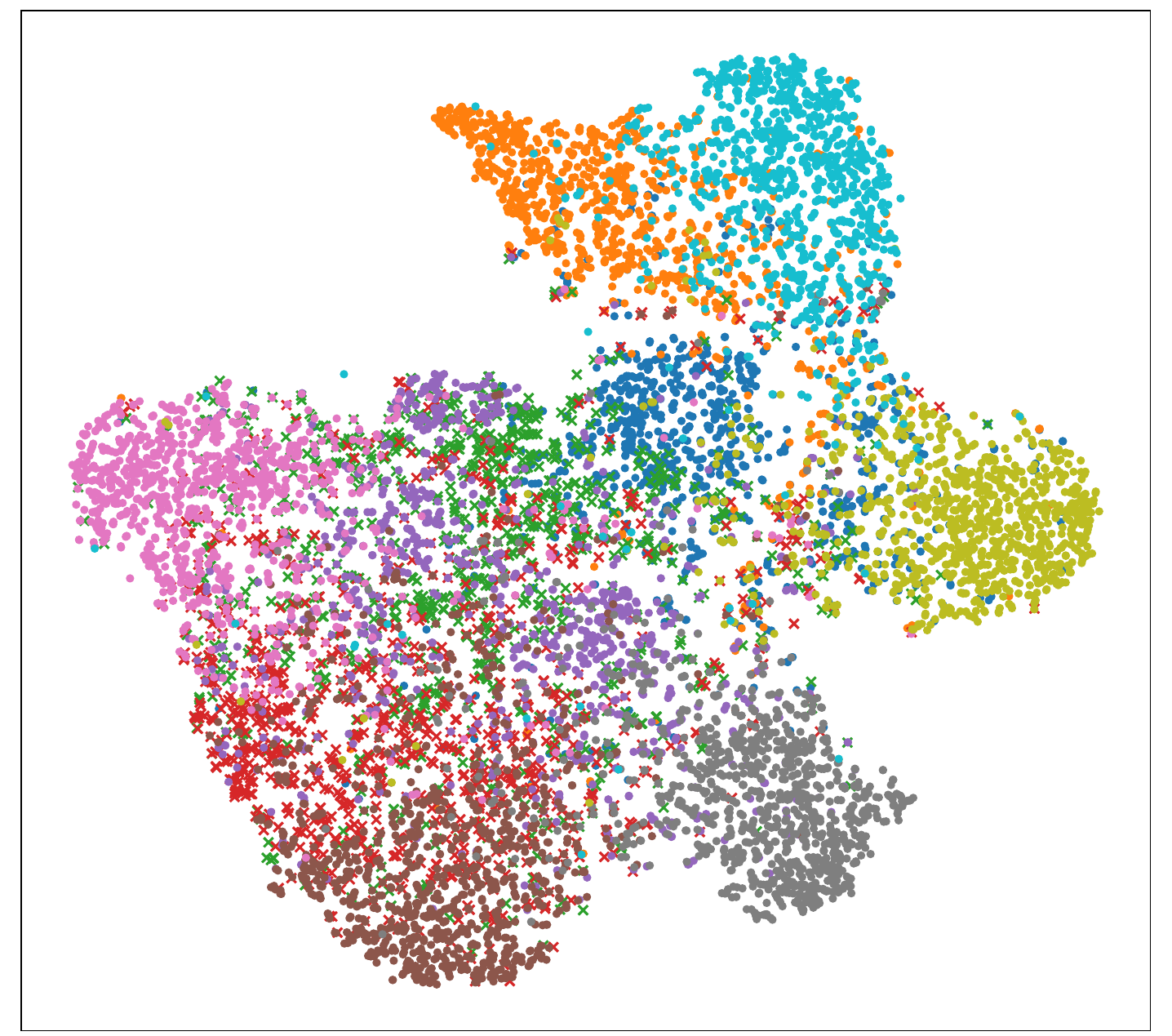}\\
Features from a model trained with the TLA loss

\includegraphics[width=0.7\linewidth]{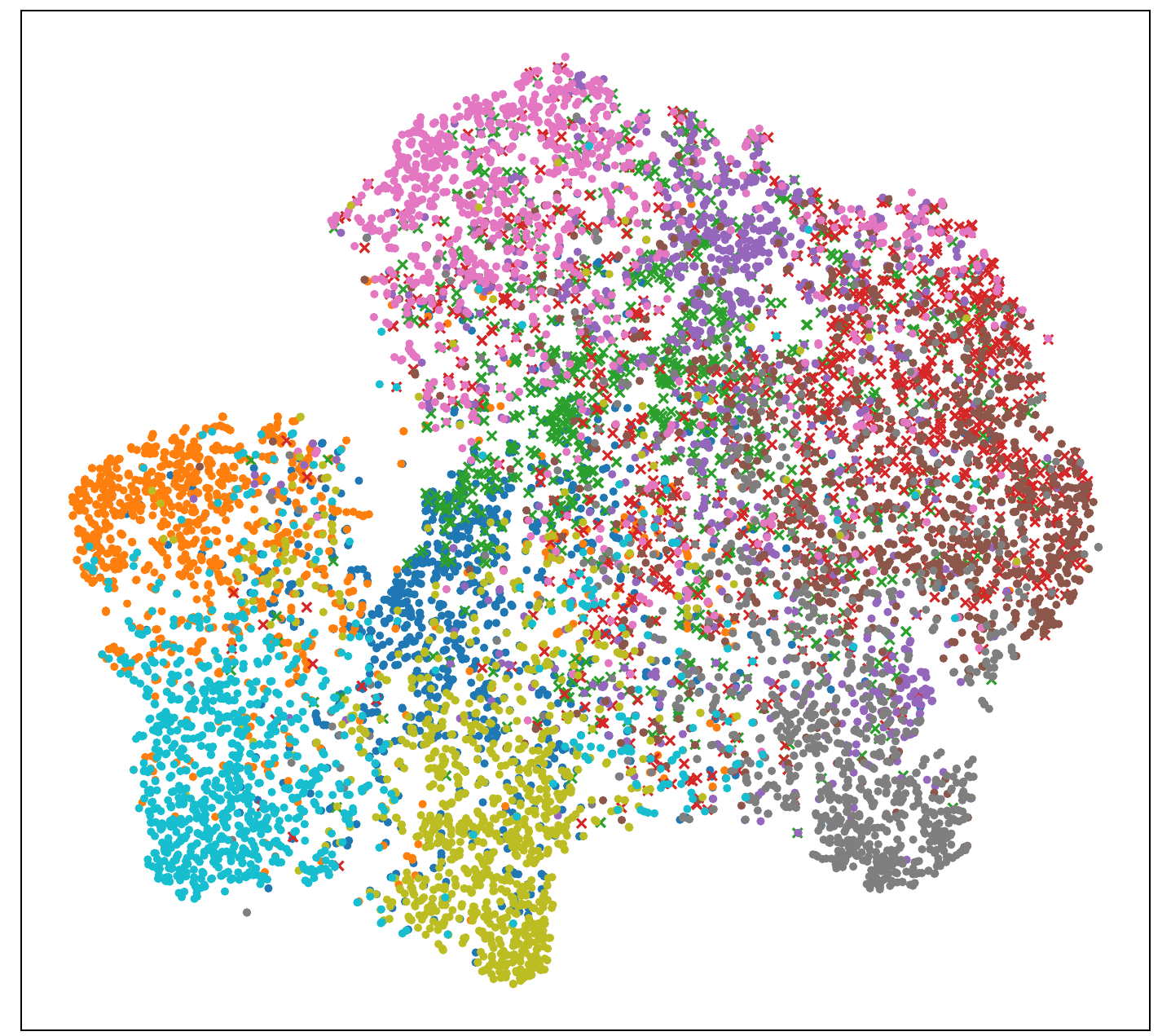}\\
Features from a model trained with the TWCE loss
\end{multicols}
\vspace{-0.15in}
\caption{Full comparison via t-SNE between the TLA and TWCE losses for CIFAR10 step imbalance with $\rho=0.01$.}\label{fig:robust_loss_full_tsne}
\vspace{-0.1in}
\end{figure*}

Note that $\pi^{t_i,\maxx,1}$ can be only one of the $K$ values as it has $K$ coordinates. Thus, the maximum value of $P$ is $K$, which implies that $\pi^*$ is included in a regular $K$-side polygon where each side has $2 \sqrt{2} \alpha$ length. This results in $\| \pi^* - \pi^{t} \| \le 2 \sqrt{2} \alpha K$ for $t>t_n$ since $\| \pi^* - \pi^{t} \|$ cannot be bigger than the length of the circumference of the formed regular $K$-side polygon by $\pi^{t}$ for $t>t_n$. See example for $K=3$ at Figure~\ref{fig:convergence_area_2}.

Now, we have established an upper bound on $\| \pi^* - \pi^{t} \|$. The remaining step is to bound $\|\min_{\theta} R(\pi^*,\theta)-\min_{\theta} R(\pi^{t},\theta)\|$ by the Lipschitzness. Recall that a function $f: \mathcal{Z} \mapsto \mathbb{R}$ is $L$-Lipschitz if
\begin{align}
    \|f(z_1)-f(z_2)\| \le L \|z_1 - z_2\| ~ \text{for all $z_1,z_2 \in \mathcal{Z}$.}
\end{align}

Note that a differentiable function is $L$-Lipschitz if and only if its differential is uniformly bounded by $L$~\cite[Lemma~2.6]{garrigos2023handbook}. Using this, we can see that $\min_{\theta} R(\pi,\theta)$ is $\sqrt{K}$-Lipschitz since $\|\nabla_{\pi} \min_{\theta} R(\pi^{t},\theta)\| \le \sqrt{K}$. Recall that $\nabla_{\pi} \min_{\theta} R(\pi^{t},\theta) = [P^{(e)}_{1,\theta^t},\ldots,P^{(e)}_{K,\theta^t}]$ and $0 \le P^{(e)}_{y,\theta^t} \le 1 $. Thus, for $\alpha=\epsilon$ and $t>t_n$,
\begin{align}
    &\left\|\min_{\theta} R(\pi^*,\theta)-\min_{\theta} R(\pi^{t},\theta)\right\|\\
    &\le \sqrt{K} \| \pi^*- \pi^{t}\| \le 2 \sqrt{2} K \sqrt{K} \alpha =2 \sqrt{2} K \sqrt{K} \epsilon.
\end{align}
This proves the claim of Theorem~\ref{thm:linear ascent convergence area}.

\section{Proof of Corollary \ref{thm:linear ascent convergence area2}}
\label{app:linear ascent convergence area2}
The most part of the proof is identical to the proof of Theorem~\ref{thm:linear ascent convergence area} except for the value $M$. Previously in Theorem~~\ref{thm:linear ascent convergence area}, $M=1$ implied that $\pi^{t,\maxx,1}$ could have $K$ values at most. For $M > 1$, $\pi^{t,\maxx,M}$ could have at most $\binom{K}{M}$ different values. Then, for $\alpha=\epsilon$ and $t>t_n$, it holds that
\begin{align}
    \left\|\min_{\theta} R(\pi^*,\theta)-\min_{\theta} R(\pi^t,\theta)\right\| \le 2 \sqrt{2} \binom{K}{M} \sqrt{K} \epsilon.
\end{align}

\section{Detailed experiment Setting}
\label{app:hyperparameters}

Table~\ref{tab:hyperparameter of loss} shows the hyperparameters of loss functions for different datasets. For fair experiment, these values are set to be the same as previous imbalanced data research~\cite{kini2021label}. 

\begin{table}[t]
\centering
\caption{Hyperparameters of loss functions for different datasets.}
\label{tab:hyperparameter of loss}

\begin{tabular}{| l | l | l | l | l |}
\hline
Dataset  & \multicolumn{2}{c|}{CIFAR10}   & \multicolumn{2}{c|}{CIFAR100} \\ \hline
Imbalance type     & LT             & step          & LT               & step
\\ \hline
LA ($\tau$)        & 2.25           & 2.25          & 1.375            & 0.875  
\\ \hline
VS ($\tau, \gamma$) & (1.25,0.15)   & (1.5,0.2)     & (0.75,0.05)      & (0.5,0.05)    
\\ \hline
TLA ($\tau$)       & 2.25           & 2.25          & 1.375            & 0.875       
\\ \hline
\end{tabular}
\vspace{-0.1in}
\end{table}

\section{Full T-SNE Figure}
\label{full_tsne_result}

Figure~\ref{fig:robust_loss_full_tsne} shows the full t-SNE figure for CIFAR10 with $\rho=0.01$ step-imbalance case.

\end{document}